\documentclass{article}

\usepackage{microtype}
\usepackage{graphicx}
\usepackage{subfigure}
\usepackage{booktabs} 

\usepackage{hyperref}

\usepackage[accepted]{icml2024}

\usepackage{amsmath}
\usepackage{amssymb}
\usepackage{mathtools}
\usepackage{amsthm}

\usepackage[capitalize,noabbrev]{cleveref}

\usepackage{multirow}

\theoremstyle{plain}
\newtheorem{theorem}{Theorem}[section]
\newtheorem{proposition}[theorem]{Proposition}

\theoremstyle{definition}
\newtheorem{definition}[theorem]{Definition}

\theoremstyle{remark}

\usepackage[textsize=tiny]{todonotes}

\icmltitlerunning{Less is More: on the Over-Globalizing Problem in Graph Transformers}

\begin{document}

\twocolumn[
\icmltitle{Less is More: on the Over-Globalizing Problem in Graph Transformers}




\begin{icmlauthorlist}
\icmlauthor{Yujie Xing}{bupt}
\icmlauthor{Xiao Wang}{buaa}
\icmlauthor{Yibo Li}{bupt}
\icmlauthor{Hai Huang}{bupt}
\icmlauthor{Chuan Shi}{bupt}
\end{icmlauthorlist}

\icmlaffiliation{bupt}{School of Computer Science, Beijing University of Posts and Telecommunications, Beijing, China}
\icmlaffiliation{buaa}{School of Software, Beihang University, Beijing, China}

\icmlcorrespondingauthor{Xiao Wang}{xiao\_wang@buaa.edu.cn}
\icmlcorrespondingauthor{Chuan Shi}{shichuan@bupt.edu.cn}

\icmlkeywords{Machine Learning, ICML}

\vskip 0.3in
]



\printAffiliationsAndNotice{}  

\begin{abstract}
Graph Transformer, due to its global attention mechanism, has emerged as a new tool in dealing with graph-structured data. It is well recognized that the global attention mechanism considers a wider receptive field in a fully connected graph, leading many to believe that useful information can be extracted from all the nodes. In this paper, we challenge this belief: does the globalizing property always benefit Graph Transformers? We reveal the over-globalizing problem in Graph Transformer by presenting both empirical evidence and theoretical analysis, i.e., the current attention mechanism overly focuses on those distant nodes, while the near nodes, which actually contain most of the useful information, are relatively weakened. Then we propose a novel Bi-Level Global Graph Transformer with Collaborative Training (CoBFormer), including the inter-cluster and intra-cluster Transformers, to prevent the over-globalizing problem while keeping the ability to extract valuable information from distant nodes. Moreover, the collaborative training is proposed to improve the model's generalization ability with a theoretical guarantee. Extensive experiments on various graphs well validate the effectiveness of our proposed CoBFormer. The source code is available for
reproducibility at: \href{https://github.com/null-xyj/CoBFormer}{https://github.com/null-xyj/CoBFormer}.
\end{abstract}

\section{Introduction}
Graph-structured data, an essential and prevalent form in the real world, plays a vital role in modeling object interactions, such as social networks \cite{yang2017neural}, transportation networks, and protein-protein interaction networks. Graph Neural Networks (GNNs) \cite{GCN17,GAT18,graphsage17,yu2023learning}, as representative graph machine learning methods, effectively utilize their message-passing mechanism to extract useful information and learn high-quality representations from graph data. However, GNNs face challenges with layer stacking due to over-smoothing \cite{over-smoothing18,over-smoothing19,over-smoothing20} and over-squashing \cite{over-squashing21, over-squashing22} problems, which limit their receptive fields to near neighbors. In contrast, Transformers \cite{transformer17}, with their global attention mechanism, have shown exceptional expressive capability, which makes significant strides in various fields, including natural language processing \cite{bert18} and computer vision \cite{vit21}. Incorporating Transformers into graph data presents an excellent solution to these challenges since they naturally construct a fully connected graph and adaptively learn interaction relationships with the powerful global attention mechanism.

In graph-level tasks like molecular property prediction, numerous Graph Transformers leveraging global attention have achieved remarkable success \cite{graphormer21,san21,graphtrans21,graphgps22}. This success is largely attributed to their global perception capability. Inspired by the successful applications in graph-level tasks, researchers have attempted to solve the scalability challenge posed by the $O(N^2)$ complexity of the global attention mechanism and make efforts to adapt this mechanism for node-level task, aiming at expanding the receptive field and enhancing the model's expressive ability \cite{coarformer22,ans-gt22,hsgt23,gapformer23,nodeformer22, GOAT23, sgformer23}.

Although the global attention module has been recognized as the fundamental unit of Graph Transformer, the following question remains largely unknown:
\vspace{-0.05in}
\begin{center}
    \textit{Does the globalizing property always benefit \\ Graph Transformers?}
\end{center}
\vspace{-0.05in}
Understanding the attention mechanism in Graph Transformers, particularly its globalizing property, can provide valuable guidelines and insights for the development of advanced Graph Transformers. In this study, we reveal the over-globalizing problem in Graph Transformers by presenting both empirical evidence and theoretical analysis. In particular, we empirically find that there is an inconsistency between the distribution of learned attention scores across all node pairs and the distribution of nodes that are actually informative, i.e., the global attention mechanism tends to focus on higher-order nodes, while the useful information often appears in lower-order nodes. Despite that higher-order nodes may provide additional information, the current attention mechanism overly focuses on those nodes. Theoretically, we demonstrate that an excessively expanded receptive field can diminish the effectiveness of the global attention mechanism, further implying the existence of the over-globalizing problem.

Once the weakness of the global attention mechanism in Graph Transformers is identified, another question naturally emerges: \textit{how to improve the current global attention mechanism to prevent the over-globalizing problem in Graph Transformers, while still keeping the ability to extract valuable information from high-order nodes?} Usually, one can alleviate this problem by implicitly or explicitly integrating a local module (e.g., GNNs) to complement Graph Transformers \cite{gophormer21, ans-gt22, coarformer22, GOAT23, gapformer23, sgformer23}. However, the different properties of local smoothing in GNNs and over-globalizing in Graph Transformers raise a fundamental question about which information will predominantly influence the node representation. Moreover, the prevalent approach of fusing local and global information through linear combination is inadequate and potentially leads to incorrect predictions, even in situations where using either local or global information alone could have achieved accurate predictions.

In this paper, we propose a novel Bi-Level Global Graph Transformer with Collaborative Training (CoBFormer). Specifically, we first partition the graph into distinct clusters with the METIS algorithm. Subsequently, we propose the bi-level global attention (BGA) module, which consists of an intra-cluster Transformer and an inter-cluster Transformer. This module effectively mitigates the over-globalizing problem while keeping a global receptive ability by decoupling the information within intra-clusters and between inter-clusters. To capture the graph structure information neglected by the BGA module, a Graph Convolution Network (GCN) is adopted as the local module. Finally, we propose collaborative training to integrate the information learned by the GCN and BGA modules and boost their performance. We summarize our contributions as follows:
\begin{itemize}
    \item We demonstrate a crucial phenomenon: Graph Transformers typically yield the over-globalizing problem of attention mechanism for node classification. Both the theoretical analysis and empirical evidence are provided to show that this problem will fundamentally affect Graph Transformers. Our discoveries provide a perspective that offers valuable insights into the improvement of Graph Transformers.
    \item We propose CoBFormer, a Bi-Level Global Graph Transformer with Collaborative Training, which effectively addresses the over-globalizing problem. Theoretical analysis implies that our proposed collaborative training will improve the model's generalization ability.
    \item Extensive experiments demonstrate that CoBFormer outperforms the state-of-the-art Graph Transformers and effectively solves the over-globalizing problem. 
\end{itemize}

\section{Preliminaries}
We denote a graph as $\mathcal{G}=(\mathcal{V},\mathcal{E})$, where the node set $\mathcal{V}$ contains $N$ nodes and the edge set $\mathcal{E}$ contains $E$ edges. All edges formulate an adjacency matrix $\mathbf{A} = [a_{uv}] \in \{0,1\}^{N\times N}$, where $a_{uv}=1$ if there exists an edge from node $u$ to $v$, and 0 otherwise. Graph $\mathcal{G}$ is often associated with a node feature matrix $\mathbf{X}=[\mathbf{x}_{u}] \in \mathbb{R}^{N \times d}$, where $\mathbf{x}_{u}$ is a $d$ dimensional feature vector of node $u$. The label set is denoted as $\mathcal{Y}$. Labels of nodes are represented with a label matrix $\mathbf{Y} = [\mathbf{y}_u] \in \mathbb{R}^{N \times |\mathcal{Y}|}$, where $\mathbf{y}_u$ is the one-hot label of node $u$. We use bold uppercase letters to represent matrices and bold lowercase letters to represent vectors.

\textbf{Graph Transformers.} Graph Transformers allow each node in a graph to attend to any other nodes by its powerful global attention mechanism as follows:
\begin{equation}
\label{eq:global_attn}
 \begin{gathered}
 \text{Attn}(\mathbf{H}) = \text{Softmax}\left(\frac{\mathbf{Q}\mathbf{K}^T}{\sqrt{h}}\right)\textbf{V}, \\
 \mathbf{Q}=\mathbf{HW}_Q,\mathbf{K}=\mathbf{HW}_K,\mathbf{V}=\mathbf{HW}_V,
 \end{gathered}
\end{equation}
where $\mathbf{H} \in \mathbb{R}^{N \times h}$ denotes the hidden representation matix and $h$ is the hidden representation dimension. $\mathbf{W}_Q, \mathbf{W}_K, \mathbf{W}_V \in \mathbb{R}^{h \times h}$ are trainable weights of linear projection layers. The attention score matrix is $\mathbf{\hat A} = \text{Softmax}\left(\frac{\mathbf{Q}\mathbf{K}^T}{\sqrt{h}}\right) \in \mathbb{R}^{N \times N}$, containing the attention scores of any node pairs. $\alpha_{uv}$ is the element of $\mathbf{\hat A}$, representing the attention score between node $u$ and $v$. It can be seen that Graph Transformers globally update the node representations by multiplying the attention score matrix $\mathbf{\hat A}$ with the node representation matrix $\mathbf{V}$.

\begin{figure*}[t]
\vskip -0.15in
\begin{center}
\subfigure[]{
\label{fig:CkReal}
\includegraphics[width=0.65\columnwidth]{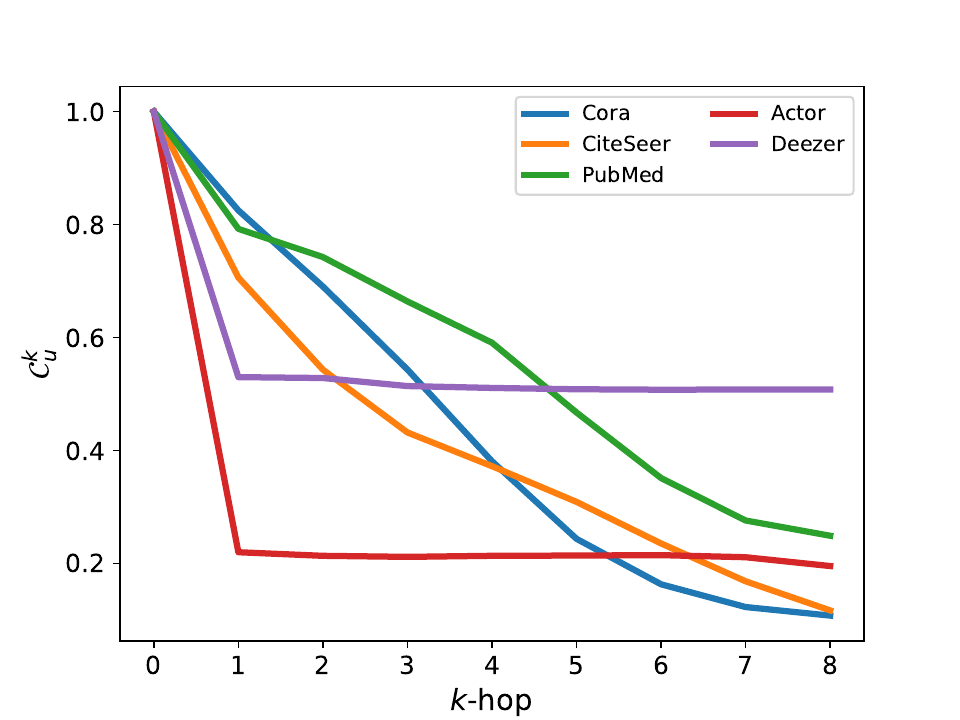}
}\subfigure[]{
\label{fig:attn_k_vt}
\includegraphics[width=0.65\columnwidth]{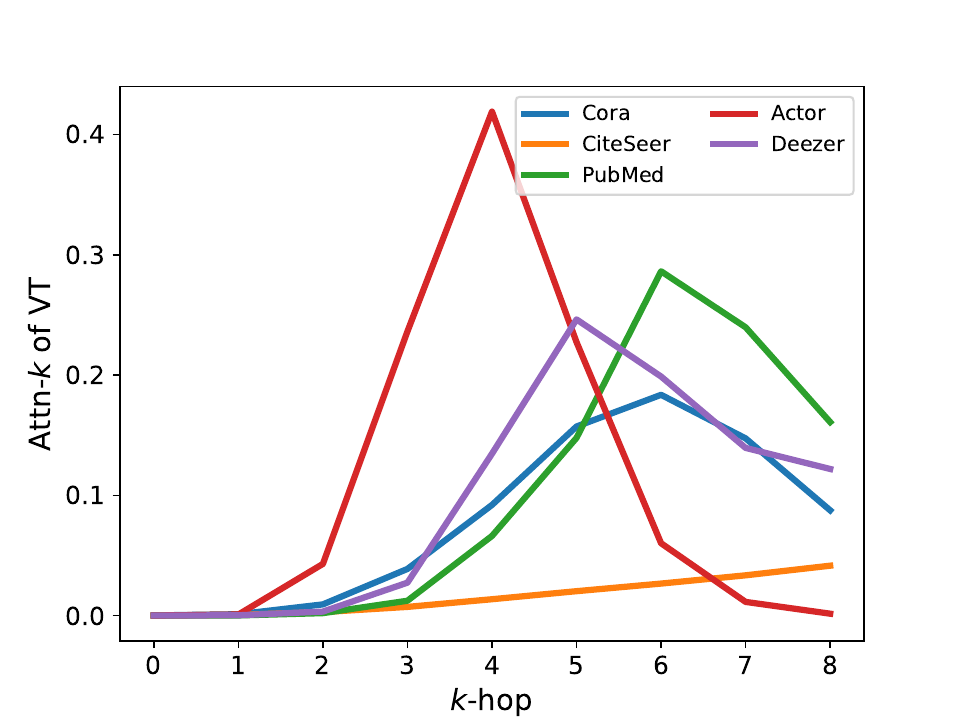}
}\subfigure[]{
\label{fig:attn_k_node}
\includegraphics[width=0.65\columnwidth]{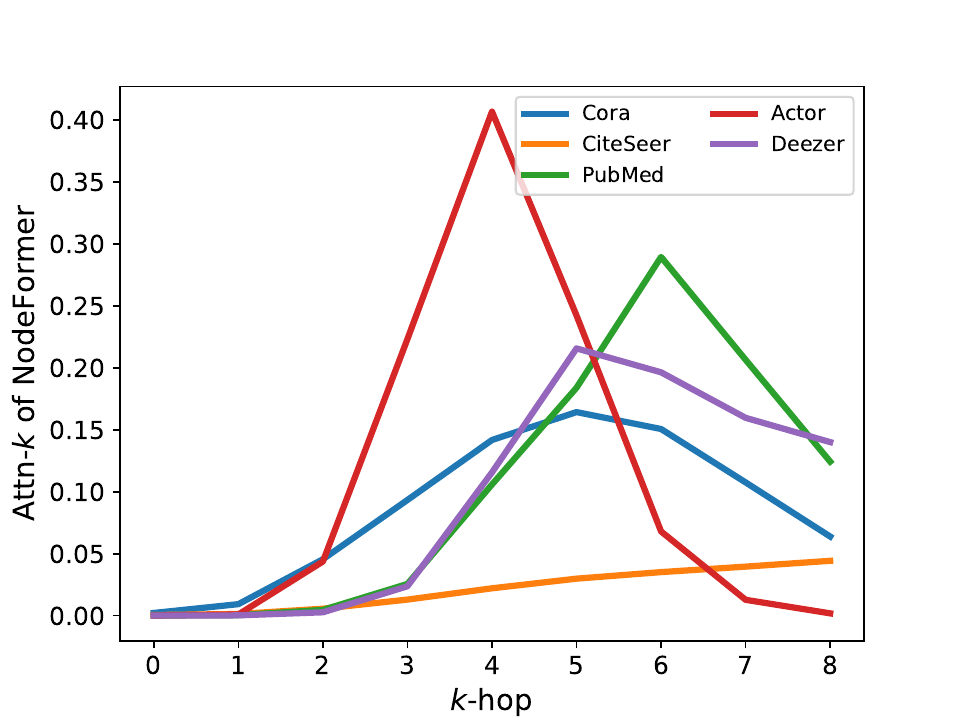}
}
\vskip -0.1in
\caption{(a) The average $\mathcal{C}_u^k$ with different $k$-hop on five real-world datasets. (b) The Attn-$k$ of Vanilla Transformer. (c) The Attn-$k$ of NodeFormer.}
\end{center}
\vskip -0.2in
\label{fig:attn_k}
\end{figure*}
\section{Over-Globalizing Problem}
\label{over-globalizing}
\textbf{Empirical observations.} In this section, we closely examine the distribution of the attention scores $\alpha_{uv}$ to investigate what information the attention mechanism captures. Specifically, we define the proportion of the $k$-th hop neighbors sharing the same label with node $u$ as follows:
\begin{equation}
    \mathcal{C}_u^k = \frac{|v \in \mathcal{N}^k(u):\mathbf{y}_u = \mathbf{y}_v|}{|\mathcal{N}^k(u)|},
\end{equation}
where $\mathcal{N}^k(u)$ denotes the $k$-th hop neighbors of node $u$. Larger $\mathcal{C}_u^k$ indicates a higher proportion of useful nodes in the $k$-th hop neighbors. Then we denote the average attention scores allocated to the $k$-th hop neighbors as Attn-$k$, which is formulated as:
\begin{equation}
    \text{Attn-}k = \mathbb{E}_{u \in \mathcal{V}} \sum_{v \in \mathcal{N}^k(u)} \alpha_{uv}.
\end{equation}
Larger Attn-$k$ implies that the model pays more attention to the $k$-th hop information. 

We present the changes of the average $\mathcal{C}_u^k$ across three homophilic graphs (Cora, CiteSeer and PubMed) and two heterophilic graphs (Actor and Deezer) in \cref{fig:CkReal}. We can observe that: (1) For homophilic graphs, $\mathcal{C}_u^k$ will gradually decrease as the $k$ increases. (2) For heterophilic graphs, $\mathcal{C}_u^k$ will rapidly decrease when $k=1$ and then remains nearly unchanged. This demonstrates that homophilic graphs benefit more from the local structure for node classification, while heterophilic graphs gain more information from the global receptive field. Then we visualize the Attn-$k$ of Vanilla Transformer (VT) and NodeFormer \cite{nodeformer22} to check whether the trend of Attn-$k$ is consistent with \cref{fig:CkReal}. To achieve this visualization, we take the average attention score across each head and plot the results for the first layer. It is worth noting that the subsequent layers will exhibit a similar trend. As can be seen in \cref{fig:attn_k_vt,fig:attn_k_node}, surprisingly, we find that the majority of the attention scores are actually allocated to distant higher-order neighbors, regardless of whether the graphs are homophilic or heterophilic. We identify this phenomenon as the over-globalizing problem in Graph Transformers, underscoring the limitations of the global attention mechanism.

\textbf{Theoretical analysis.} Here we further theoretically explore the impact of the over-globalizing problem in Graph Transformers. Ideally, Graph Transformers would allocate higher attention scores to nodes with similar embeddings, thereby implicitly learning a graph structure that ensures the smoothness of embeddings among adjacent nodes. Consequently, $\| \mathbf{Z} - \mathbf{\hat AZ} \|_F$ would be relatively small \cite{graphdenoise13,graphdenoise16}. Here $\mathbf{Z}$ symbolizes the node embeddings. So we employ $\| \mathbf{Z} - \mathbf{\hat AZ} \|_F$ to evaluate the smoothness of the embeddings among adjacent nodes learned by Graph Transformers. A smaller $\| \mathbf{Z} - \mathbf{\hat AZ} \|_F$ indicates a better smoothness, suggesting that Graph Transformers can effectively recognize useful nodes and aggregate information from them, achieving better node classification performance.

Then we investigate the factors influencing $\| \mathbf{Z} - \mathbf{\hat AZ} \|_F$. Before that, we denote $\mathcal{C}_u$ as the proportion of nodes belonging to the same class in the reachable set of node $u$. If the reachable set of node $u$ is the $K$-hop neighbors, then $\mathcal{C}_u$ can be formulated as follows:
\begin{equation}
    \mathcal{C}_u =\frac{\sum^K_{k=0}\mathcal{C}_u^k|\mathcal{N}^k(u)|}{\sum^K_{k=0}|\mathcal{N}^k(u)|}.
\end{equation}

Now we can establish the connection between $\| \mathbf{Z} - \mathbf{\hat AZ} \|_F$, $\alpha_{uv}$ and $\mathcal{C}_u$ as follows:
\begin{theorem}
\label{thm:denoise2attn}
For a given node $u$ and a well-trained Graph Transformer, let $\eta_u=\mathbb{E}_{v \in \mathcal{V},\mathbf{y}_u = \mathbf{y}_v}\exp(\frac{\mathbf{q}_u\mathbf{k}_v^T}{\sqrt{d}})$, $\gamma_u=\mathbb{E}_{v \in \mathcal{V},\mathbf{y}_u \neq \mathbf{y}_v}\exp(\frac{\mathbf{q}_u\mathbf{k}_v^T}{\sqrt{d}})$. Then, we have:
\begin{equation}
\label{eq:denoise2attn}
\begin{aligned}
    \| \mathbf{Z} - \mathbf{\hat AZ} \|_F &\le \sqrt{2}L\sum_{u \in \mathcal{V}}\sum_{v \in \mathcal{V},\mathbf{y}_u \neq \mathbf{y}_v} \alpha_{uv} \\
    &= \sqrt{2}L\sum_{u \in \mathcal{V}}\frac{1}{1+\frac{\mathcal{C}_u}{1-\mathcal{C}_u}\frac{\eta_u}{\gamma_u}}.
\end{aligned}
\end{equation}
where $L$ is a Lipschitz constant. 
\end{theorem}
The proofs are given in \cref{appendix:proof1}. \cref{thm:denoise2attn} indicates that $ \| \mathbf{Z} - \mathbf{\hat AZ} \|_F$ is bounded by the sum of the attention scores of node pairs with different labels and negatively correlated with $\mathcal{C}_u$, since $\eta_u$ and $\gamma_u$ are constants for a given Graph Transformer. Then we further study the variations of $\mathcal{C}_u$ in \cref{thm:same-label-rate}.

\begin{theorem}
\label{thm:same-label-rate}
To analyze the impact of $k$ on $\mathcal{C}_u^k$, we assume that each node has an equal probability $\frac{1}{|\mathcal{Y}|}$ of belonging to any given class. Given the edge homophily $\rho=\frac{|(u,v)\in\mathcal{E}:\mathbf{y}_u=\mathbf{y}_v|}{|\mathcal{E}|}$, $\mathcal{C}_u^k$ can be recursively defined as:
\begin{equation}
 \mathcal{C}_u^k = \begin{cases}
 1, & if \ k=0 \\
 \rho, & if \ k=1 \\
 \frac{1+|\mathcal{Y}|\rho\mathcal{C}_u^{k-1}-\rho-\mathcal{C}_u^{k-1}}{|\mathcal{Y}|-1}. & if \ k=2,3,\cdots
 \end{cases}
\end{equation}
And $\mathcal{C}_u^k$ possesses the following properties:
\begin{equation}
\begin{cases}
 \mathcal{C}_u^\infty = \frac{1}{|\mathcal{Y}|} \\
 \mathcal{C}_u^k \ge \mathcal{C}_u^{k+1}, & if \ \rho \ge \frac{1}{|\mathcal{Y}|}, \ k=0,1\cdots \\
 \mathcal{C}_u^{2k}>\mathcal{C}_u^{2(k+1)}, & if \ \rho < \frac{1}{|\mathcal{Y}|}, \ k=0,1\cdots \\
 \mathcal{C}_u^{2k+1}<\mathcal{C}_u^{2(k+1)+1}, & if \ \rho < \frac{1}{|\mathcal{Y}|}. \ k=0,1\cdots \\
\end{cases}
\end{equation}
\end{theorem}

\begin{figure}[ht]
\vskip -0.1in
\begin{center}
\centerline{\includegraphics[width=0.95\columnwidth]{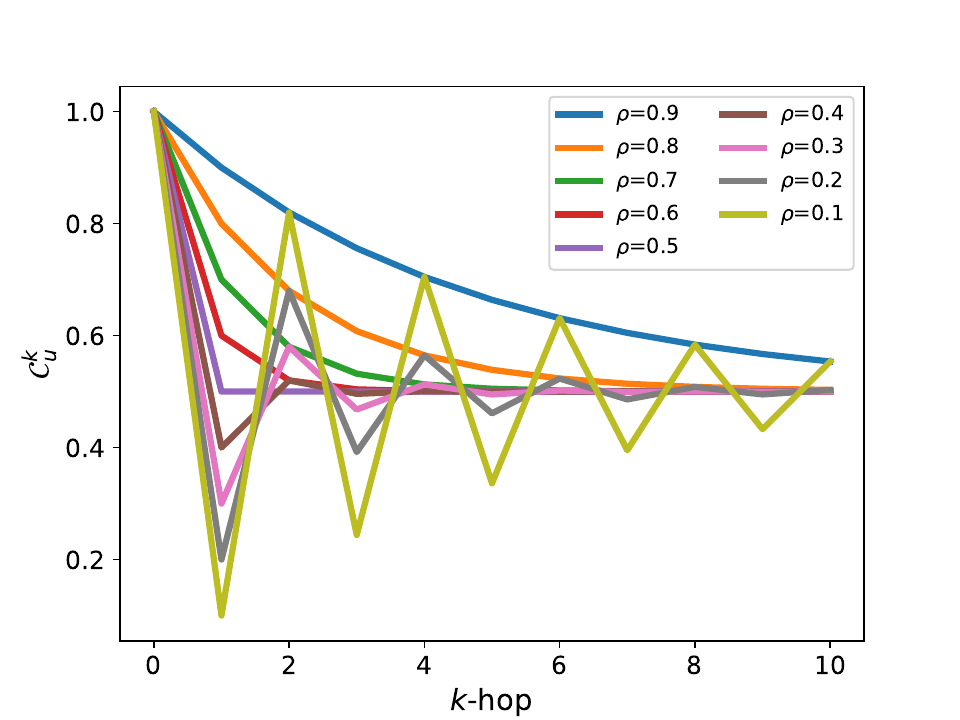}}
\caption{The variations of $\mathcal{C}_u^k$ under various $\rho$ settings for a scenario involving only two classes with uniformly distributed labels.}
\label{fig:Ck}
\end{center}
\end{figure}

We provide the proof in \cref{appendix:proof2}. \cref{thm:same-label-rate} indicates that in homophilic graphs, where $\rho$ is relatively large, $\mathcal{C}_u^k$ will gradually decrease and converge to $\frac{1}{\mathcal{|Y|}}$, as the $k$ increases. However, in heterophilic graphs, where $\rho$ is relatively small, it will fluctuate around $\frac{1}{\mathcal{|Y|}}$ and eventually converge to $\frac{1}{\mathcal{|Y|}}$. Combining with \cref{thm:denoise2attn}, we find that in homophilic graphs, as the receptive field expands, the gradually decreased $\mathcal{C}_u^k$ will lead to a reduced $\mathcal{C}_u$ and a larger $\| \mathbf{Z} - \mathbf{\hat AZ} \|_F$, implying that an over-expanded receptive field adversely affects the global attention. Conversely, in heterophilic graphs, global attention brings in additional information that cannot be captured within the local neighborhood. Based on \cref{thm:same-label-rate}, we visualize the theoretical variations of $\mathcal{C}_u^k$ in \cref{fig:Ck}. Compared with \cref{fig:CkReal}, we can find that our theories align well with real-world scenarios. More visualization results of theoretical scenarios are provided in \cref{appendix:visual}.

\textbf{Experimental analysis.} Inspired by \cref{thm:denoise2attn}, we define the Attention Signal/Noise Ratio (Attn-SNR) as the metric to quantify the ability of Graph Transformers to distinguish useful nodes as follows:
\begin{definition} 
\label{def:Attn-SNR}
The Attention Signal/Noise Ratio (Attn-SNR) is:
\vspace{-0.05in}
\begin{equation}
 \text{Attn-SNR} = 10\lg\left(\frac{\sum_{\mathbf{y}_u = \mathbf{y}_v}\alpha_{uv}}{\sum_{\mathbf{y}_u \neq \mathbf{y}_v}\alpha_{uv}}\right).
\end{equation}
\end{definition}
\vspace{-0.05in}
For a given Graph Transformer, a smaller Attn-SNR usually implies that the attention mechanism pays more attention to nodes with different labels, which may be caused by the over-globalizing problem. We evaluate Vanilla Transformer and NodeFormer utilizing Attn-SNR and accuracy on Cora and Citeseer. Furthermore, we deliberately improve the Attn-SNR of Vanilla Transformer by doubling the attention scores between nodes sharing the same label, and report its performance. The results are presented in \cref{noise-table}, indicating that: (1) Vanilla Transformer (VT) typically shows the least Attn-SNR, resulting in the poorest performance. NodeFormer (NF) exhibits a higher Attn-SNR and achieves superior performance. (2) Remarkably, the Denoised Vanilla Transformer (VT-D), artificially directed to achieve higher Attn-SNR, demonstrates better performance than Vanilla Transformer. This is because the over-globalizing problem can be alleviated by doubling the attention scores between nodes with the same label, which are more likely to appear in the local neighborhood, thereby enhancing the model's classification capability.
\begin{table}[ht]
\caption{The Attn-SNR and testing accuracy of different models.}
\label{noise-table}
\begin{center}
\begin{small}
\begin{tabular}{l|c|ccc}
\toprule
Dataset & Metric & VT & NF & VT-D \\
\midrule
\multirow{2}{*}{Cora} & Attn-SNR & -6.97 & 0.43 & 12.05 \\
& Accuracy & 55.18 & 80.20 & 82.12 \\ 
\midrule
\multirow{2}{*}{CiteSeer} & Attn-SNR & -7.19 & -5.09 & 8.72 \\  
& Accuracy & 50.72 & 71.50 & 61.80 \\ 
\bottomrule
\end{tabular}
\end{small}
\end{center}
\vskip -0.2in
\end{table}

\begin{figure*}[t]
\begin{center}
\centerline{\includegraphics[width=1.8\columnwidth]{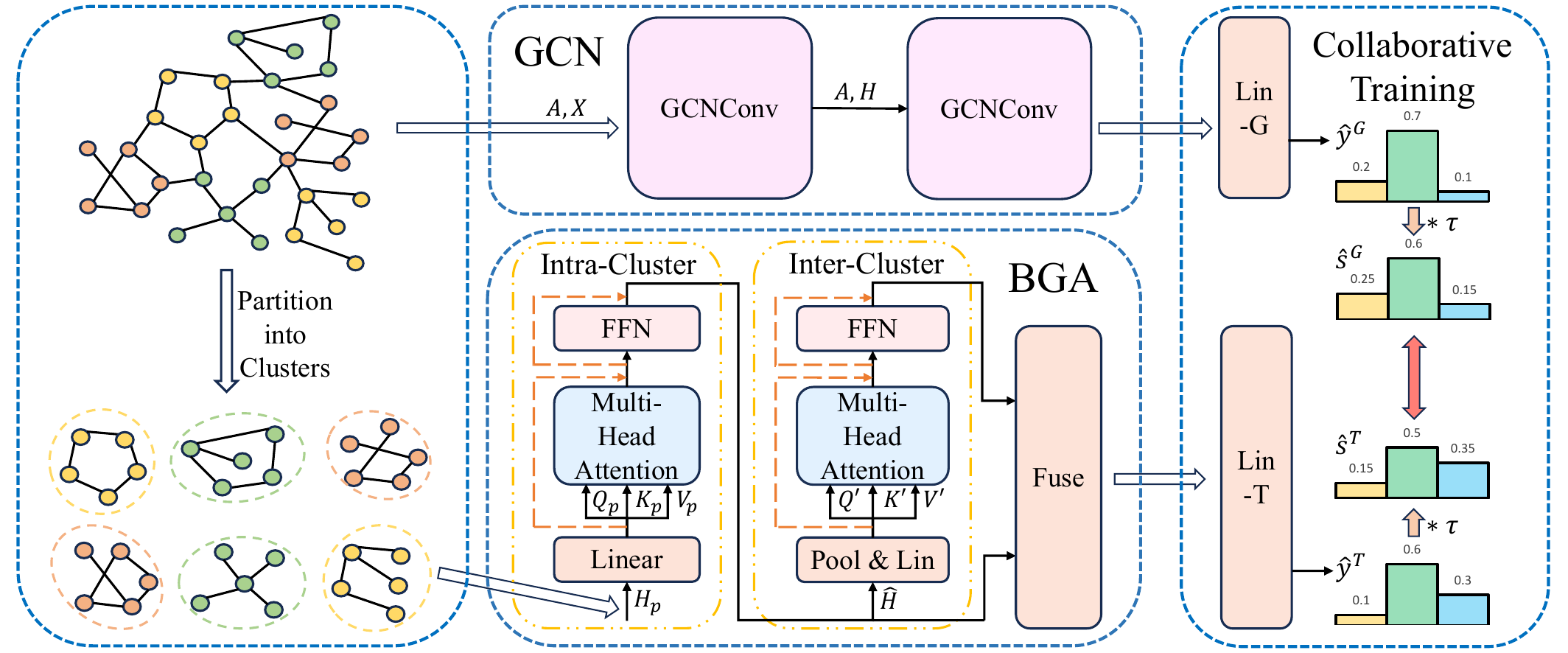}}
\caption{The overall framework of our proposed CoBFormer.}
\label{fig:model}
\end{center}
\vskip -0.3in
\end{figure*}
\section{The Proposed Method}
In this section, we introduce the Bi-Level Global Graph Transformer with Collaborative Training (CoBFormer). An overview of CoBFormer is shown in \cref{fig:model}. 
Specifically, we first use the METIS algorithm \cite{metis98} to partition the graph into different clusters. Then we propose a novel bi-level global attention (BGA) module, which can decouple the information within intra-clusters and between inter-clusters by an intra-cluster Transformer and an inter-cluster Transformer. Additionally, we incorporate a GCN as the local module to learn the graph structural information. Finally, we propose collaborative training to integrate the information obtained by the GCN and BGA modules and theoretically prove that the generalization ability can be improved with our proposed collaborative training.

\subsection{Bi-Level Global Attention Module} 

Traditional Graph Transformers utilize the global attention mechanism to capture information between any node pairs, causing the over-globalizing problem. Therefore, we need to guarantee that local information can be captured, so as to alleviate the problem. To achieve this goal, we first partition the graph into $P$ non-overlapping clusters using METIS \cite{metis98}.
We denote the set of clusters as $\mathcal{P}=\{\mathcal{G}_i\}$, where $\mathcal{G}_i=\{\mathcal{V}_i,\mathcal{E}_i\}$ represents a subgraph of $\mathcal{G}$, satisfying $\bigcup \mathcal{G}_i=\mathcal{G},\bigcap \mathcal{G}_i=\varnothing$. 

 The local information is usually within each cluster, so we employ an intra-cluster Transformer. The node features in cluster $p$ are represented as $\mathbf{X}_p \in \mathbb{R}^{\frac{N}{P} \times d}$. We apply an MLP to project the original node features into a latent space as $\mathbf{H}_p^1=\text{MLP}(\mathbf{X}_p)$. Subsequently, the hidden representations $\mathbf{H}_p^1 \in \mathbb{R}^{\frac{N}{P} \times h}$ are fed into the intra-cluster Transformer to learn the updated hidden representations $\mathbf{\hat H}^k_p$:
\begin{equation}
 \begin{gathered}
 \mathbf{\hat H}^k_p = \text{FFN}\left(\text{Softmax}\left(\frac{\mathbf{Q}_p\mathbf{K}_p^T}{\sqrt{h}}\right)\mathbf{V}_p\right), \\
 \mathbf{Q}_p=\mathbf{H}_p^k\mathbf{W}^k_Q,\mathbf{K}_p=\mathbf{H}_p^k\mathbf{W}^k_K,\mathbf{V}_p=\mathbf{H}_p^k\mathbf{W}^k_V,
 \end{gathered}
\end{equation}
where $\mathbf{W}_{Q}^k, \mathbf{W}_{K}^k$ and $\mathbf{W}_{V}^k \in \mathbb{R}^{h \times h}$ are trainable weights of the linear projection layers in the $k$-th intra-cluster Transformer, and FFN represents a Feed-Forward Neural Network. In each attention block and FFN block, a residual connection \cite{residual16} and a layer normalization \cite{layernorm} are applied.

Subsequently, we apply mean pooling to $\mathbf{\hat H}^k_p$ to obtain the cluster representations $\mathbf{P}^k \in \mathbb{R}^{P \times h}$. The $p$-th row of $\mathbf{P}^k$, represented as $\mathbf{p}_p^k$, is calculated by $\mathbf{p}_p^k=\text{MEAN}(\mathbf{\hat H}^k_p) \in \mathbb{R}^{1\times h}$. Following this, the cluster representations $\mathbf{P}^k$ are fed into the inter-cluster Transformer:
\begin{equation}
\label{eq:inter-trans}
 \begin{gathered}
 \mathbf{\hat P}^k = \text{FFN}\left(\text{Softmax}\left(\frac{\mathbf{Q'}\mathbf{K'}^T}{\sqrt{h}}\right)\mathbf{V'}\right), \\
 \mathbf{Q'}=\mathbf{P}^k\mathbf{W}^k_{Q'},\mathbf{K}'=\mathbf{P}^k\mathbf{W}^k_{K'},\mathbf{V}'=\mathbf{P}^k\mathbf{W}^k_{V'},
 \end{gathered}
\end{equation}

where $\mathbf{W}_{Q'}^k, \mathbf{W}_{K'}^k$ and $\mathbf{W}_{V'}^k \in \mathbb{R}^{h \times h}$ are trainable weights of the linear projection layers in the $k$-th inter-cluster Transformer. Although the inter-cluster Transformer learns the attentions between different clusters, it can approximate the global attention mechanism in \cref{eq:global_attn} and effectively learn the global information as in \cref{prop:approximate-global}.
\begin{proposition}
\label{prop:approximate-global}
Given $u \in \mathcal{V}_p, v \in \mathcal{V}_q$,  along with a well-trained inter-cluster attention score matrix $\mathbf{\dot A} \in \mathbb{R}^{P \times P}$. Let $\dot \alpha_{pq}$ represent the attention score between clusters $p$ and $q$. Then the approximate attention score between node $u$ and $v$ can be expressed as $\hat \alpha_{uv}=\frac{\dot \alpha_{pq}}{|\mathcal{V}_q|}$.
\end{proposition}
The proof can be found in \cref{appendix:proof4}. It can be seen that the attention score $\alpha_{uv}$ of the global attention mechanism can be approximately represented as $\frac{\dot \alpha_{pq}}{|\mathcal{V}_q|}$, which can be calculated by our inter-cluster Transformer. 

Now with both $\mathbf{\hat H}^k_p$ and  $\mathbf{\hat P}^k$, we concatenate the node representation with its corresponding cluster representation and calculate the output node representations $\mathbf{H}_p^{k+1}$ with a fusion linear layer parameterized by $\mathbf{W}_f$:
\begin{equation}
\label{eq:concat}
 \mathbf{H}_p^{k+1} = \left(\mathbf{\hat H}^k_p\|\mathbf{1}^{\frac{N}{P}}\mathbf{\hat p}_p^T\right)\mathbf{W}_f,
\end{equation}
here $\|$ indicates the concatenation operation, $\mathbf{1}^{\frac{N}{P}}$ is an all-one column vector with a dimensional of $\frac{N}{P}$ and $\mathbf{\hat p}_p$ is the hidden representation of cluster $p$.
By decoupling the information within intra-clusters and between inter-clusters, our BGA module can alleviate the over-globalizing problem while preserving the expressive ability of global attention. Concurrently, our method achieves significant computational and memory efficiency by focusing on intra-cluster and inter-cluster attention. The time and space complexity of our BGA module are $O(\frac{N^2}{P}+P^2)$, reaching $O(N^{\frac{4}{3}})$ under the optimal condition. Note that the efficiency of our BGA module can be further improved by linear attention techniques \cite{performer20, linformer20, sgformer23}.

\subsection{Collaborative Training}
\label{co-train}
With the proposed BGA module capturing intra-cluster and inter-cluser information, we further introduce a GCN as the local module to supplement the graph structure information ignored by the BGA module. Instead of directly employing a linear combination of information from the local module and the global attention module for node classification, we propose a collaborative training approach for the GCN and BGA modules. We denote the labeled node set as $\mathcal{V_L}$ and the unlabeled node set as $\mathcal{V_U}$, with $\mathbf{L}$ and $\mathbf{U}$ representing their respective label matrix. In previous works, a model is trained by predicting the label distribution of the labeled nodes with a cross-entropy loss. However, this method does not guarantee satisfactory performance on unlabeled nodes. 

Here, we employ two linear layers, Lin-G and Lin-T, to map the outputs of the GCN and BGA modules onto the label space:
\begin{equation}
\begin{aligned}
 \mathbf{\hat Z}^G &= \text{Lin-G}(\text{GCN}(\mathbf{A}, \mathbf{X})), \\
 \mathbf{\hat Z}^T &= \text{Lin-T}(\text{BGA}(\mathbf{X},\mathcal{P})).
\end{aligned}
\end{equation}
Then we use the SoftMax function to calculate the predicted labels and soft labels \cite{kd15}:
\begin{equation}
\begin{gathered}
 \mathbf{\hat Y}^G = \text{SoftMax}(\mathbf{\hat Z}^G),\ \mathbf{\hat Y}^T = \text{SoftMax}(\mathbf{\hat Z}^T), \\
 \mathbf{\hat S}^G = \text{SoftMax}(\mathbf{\hat Z}^G*\tau),\ \mathbf{\hat S}^T = \text{SoftMax}(\mathbf{\hat Z}^T*\tau),
\end{gathered}
\end{equation}
where $\tau$ is a temperature coefficient used to control the smoothness of the soft labels. The objective function can be formulated as:
\begin{equation}
\label{eq:loss}
\begin{aligned}
 \mathcal{L}_{ce} &= -\left(\mathbb{E}_{\mathbf{y}_u,u\in\mathcal{V_L}}\log(\mathbf{\hat y}_u^G) + \mathbb{E}_{\mathbf{y}_u,u\in\mathcal{V_L}}\log(\mathbf{\hat y}_u^T)\right), \\
 \mathcal{L}_{co} &= -\left(\mathbb{E}_{\mathbf{\hat s}^G_u,u\in\mathcal{V_U}}\log(\mathbf{\hat s}_u^T) + \mathbb{E}_{\mathbf{\hat s}^T_u,u\in\mathcal{V_U}}\log(\mathbf{\hat s}_u^G)\right), \\
 \mathcal{L} &= \alpha * \mathcal{L}_{ce} + (1-\alpha) * \mathcal{L}_{co}.
\end{aligned}
\end{equation}
where $\mathbf{y}_u$ represents the true label of node $u$. $\mathbf{\hat y}_u^G$ and $\mathbf{\hat y}_u^T$ are the predicted labels of node $u$ by the GCN and BGA modules, respectively. $\mathbf{\hat s}_u^G$ and $\mathbf{\hat s}_u^T$ denote the soft labels generated by each. $\mathcal{L}_{ce}$ is the cross-entropy loss, a standard choice for classification tasks. $\mathcal{L}_{co}$ is designed to encourage mutual supervision between the GCN and BGA modules. The parameter $\alpha$ is used to balance the contributions of $\mathcal{L}_{ce}$ and $\mathcal{L}_{co}$.

Now we prove that our proposed collaborative training can improve the generalization ability of our GCN module and BGA module, thereby achieving better classification performance.
\begin{theorem}
\label{thm:co-train}
Consider $P(\mathbf{L},\mathbf{U})$ as the true label distribution, $P_G(\mathbf{L},\mathbf{U})$ as the predicted label distribution by the GCN, and $P_T(\mathbf{L},\mathbf{U})$ as the predicted label distribution by the BGA module. The following relations hold:
\begin{equation}
\label{eq:co-train}
\begin{aligned}
\mathbb{E}_{P(\mathbf{L},\mathbf{U})}\log P_G(\mathbf{L},\mathbf{U}) =
&\mathbb{E}_{P(\mathbf{L})}\log P_G(\mathbf{L}) +\\
&\mathbb{E}_{P_T(\mathbf{U}|\mathbf{L})}\log P_G(\mathbf{U}|\mathbf{L}) -\\ 
&\text{\rm{KL}}( P_T(\mathbf{U}|\mathbf{L}) \| P(\mathbf{U}|\mathbf{L})), \\
\mathbb{E}_{P(\mathbf{L},\mathbf{U})}\log P_T(\mathbf{L},\mathbf{U}) =
&\mathbb{E}_{P(\mathbf{L})}\log P_T(\mathbf{L}) +\\
&\mathbb{E}_{P_G(\mathbf{U}|\mathbf{L})}\log P_T(\mathbf{U}|\mathbf{L}) -\\ 
&\text{\rm{KL}}( P_G(\mathbf{U}|\mathbf{L}) \| P(\mathbf{U}|\mathbf{L})), \\
\end{aligned}
\end{equation}
where $\text{KL}(\cdot\|\cdot)$ is the Kullback-Leibler divergence.
\end{theorem}
The proof is given in \cref{appendix:proof3}. 
Taking the training process of the GCN module as an example, $\mathbb{E}_{P(\mathbf{L},\mathbf{U})}\log P_G(\mathbf{L},\mathbf{U})$ is the cross entropy between $P(\mathbf{L},\mathbf{U})$ and $P_G(\mathbf{L},\mathbf{U})$. We aim to maximize it so that our model can achieve the best performance on labeled nodes and unlabeled nodes. However, it cannot be maximized directly since the label distribution of unlabeled nodes is unknown. \cref{thm:co-train} suggests that $\mathbb{E}_{P(\mathbf{L},\mathbf{U})}\log P_G(\mathbf{L},\mathbf{U})$ can be decomposed into three terms. The first term is the cross entropy between $P(\mathbf{L})$ and $P_G(\mathbf{L})$, which can be maximized by optimizing the $\mathcal{L}_{ce}$. It will ensure a good performance on labeled nodes. The second term is the cross entropy between $P_T(\mathbf{U}|\mathbf{L})$ and $P_G(\mathbf{U}|\mathbf{L})$, which can be maximized by optimizing the $\mathcal{L}_{co}$. This term indicates that we can further improve the performance of the GCN module on unlabeled nodes by collaboratively training with the BGA module. Note that the third term is the Kullback-Leibler divergence between $P_T(\mathbf{U}|\mathbf{L})$ and $P(\mathbf{U}|\mathbf{L})$, which is a constant when we optimize the GCN module and this value will gradually decrease during the optimization process of the BGA module. Therefore, the performance of the GCN module can be improved by the loss in \cref{eq:loss}. Similarly, the performance of the BGA module can be improved.

\begin{table*}[ht]
\vskip -0.1in
\caption{Quantitative results ($\%\pm\sigma$) on node classification.}
\vskip 0.05in
\label{main-table}
\begin{center}
\begin{small}
\begin{tabular}{l|c|ccccc|cc}
\toprule
Dataset & Metric & GCN & GAT & NodeFormer & NAGphormer & SGFormer & CoB-G & CoB-T \\
\midrule
\multirow{2}{*}{Cora} & Mi-F1 & 81.44 ± 0.78 & 81.88 ± 0.99 & 80.30 ± 0.66 & 79.62 ± 0.25 & 81.48 ± 0.94 & \underline{84.96 ± 0.34} & \textbf{85.28 ± 0.16} \\
& Ma-F1 & 80.65 ± 0.91 & 80.56 ± 0.55 & 79.12 ± 0.66 & 78.78 ± 0.57 & 79.28 ± 0.49 & \underline{83.52 ± 0.15} & \textbf{84.10 ± 0.28} \\
\midrule
\multirow{2}{*}{CiteSeer} & Mi-F1 & 71.84 ± 0.22 & 72.26 ± 0.97 & 71.58 ± 1.74 & 67.46 ± 1.33 & 71.96 ± 0.13 & \textbf{74.68 ± 0.33} & \underline{74.52 ± 0.48} \\
& Ma-F1 & 68.69 ± 0.38 & 65.67 ± 2.28 & 67.28 ± 1.87 & 64.47 ± 1.58 & 68.49 ± 0.65 & \underline{69.73 ± 0.45} & \textbf{69.82 ± 0.55} \\
\midrule
\multirow{2}{*}{PubMed} & Mi-F1 & 79.26 ± 0.23 & 78.46 ± 0.22 &  78.96 ± 2.71  & 77.36 ± 0.96 & 78.04 ± 0.41 & \underline{80.52 ± 0.25} & \textbf{81.42 ± 0.53} \\
& Ma-F1 & 79.02 ± 0.19 & 77.82 ± 0.22 & 78.14 ± 2.51 & 76.76 ± 0.91 & 77.86 ± 0.32 & \underline{80.02 ± 0.28} & \textbf{81.04 ± 0.49} \\
\midrule
\multirow{2}{*}{Actor} & Mi-F1 & 30.97 ± 1.21 & 30.63 ± 0.68 &  35.42 ± 1.37  & 34.83 ± 0.95 & \textbf{37.72 ± 1.00} & 31.05 ± 1.02 & \underline{37.41 ± 0.36} \\
& Ma-F1 & 26.66 ± 0.82 & 20.73 ± 1.58 &  32.37 ± 1.38  & 32.20 ± 1.11 & \underline{34.11 ± 2.78} & 27.01 ± 1.77 & \textbf{34.96 ± 0.68} \\
\midrule
\multirow{2}{*}{Deezer} & Mi-F1 & 63.10 ± 0.40 & 62.20 ± 0.41 &  63.59 ± 2.24  & 63.71 ± 0.58 & \underline{66.68 ± 0.47} & 63.76 ± 0.62 & \textbf{66.96 ± 0.37} \\
& Ma-F1 & 62.07 ± 0.31 & 60.99 ± 0.56 &  62.70 ± 2.20  & 62.06 ± 1.28 & \underline{65.22 ± 0.68} & 62.32 ± 0.94 & \textbf{65.63 ± 0.36} \\
\midrule
\multirow{2}{*}{Arxiv} & Mi-F1 & 71.99 ± 0.14 & 71.30 ± 0.11 &  67.98 ± 0.60  & 71.38 ± 0.20 & 72.50 ± 0.28 & \textbf{73.17 ± 0.18} & \underline{72.76 ± 0.11} \\
& Ma-F1 & 51.89 ± 0.19 & 48.84 ± 0.31 &  46.24 ± 0.20  & 51.38 ± 0.47 & \textbf{52.83 ± 0.31} & \underline{52.31 ± 0.40} & 51.64 ± 0.09 \\
\midrule
\multirow{2}{*}{Products} & Mi-F1 & 75.49 ± 0.24  & 76.19 ± 0.40 & 70.71 ± 0.27 & 76.41 ± 0.53 & 72.54 ± 0.80 & \underline{78.09 ± 0.16} & \textbf{78.15 ± 0.07} \\
& Ma-F1 & 37.02 ± 0.92 & 35.15 ± 0.20 & 30.09 ± 0.02 & 37.48 ± 0.38 & 33.72 ± 0.42 & \textbf{38.21 ± 0.22} & \underline{37.91 ± 0.44} \\
\bottomrule
\end{tabular}
\end{small}
\end{center}
\vskip -0.2in
\end{table*}
\section{Experiments}
\label{experiments}

\textbf{Datasets.} We select seven datasets to evaluate, including homophilic graphs, i.e., Cora, CiteSeer, Pubmed \cite{citenetwork16}, Ogbn-Arxiv, Ogbn-Products \cite{ogb20} and heterophilic graphs, i.e., Actor, Deezer \cite{heterophilic21}. For Cora, CiteSeer, PubMed, we adopt the public split offered by PyG \cite{pyg19}. For Ogbn-Arxiv and Ogbn-Products, we use the public splits in OGB \cite{ogb20}. For Actor and Deezer, we perform five random splits of the nodes into train/valid/test sets, with the ratio of 50\%:25\%:25\% \cite{heterophilic21}. The detailed statistics of the datasets can be found in \cref{appendix:dataset}.

\textbf{Baselines.} We compare our method with five baselines, including two classic GNNs: GCN \cite{GCN17} and GAT \cite{GAT18}, and three state-of-the-art Graph Transformers: NodeFormer \cite{nodeformer22}, NAGphormer \cite{NAGphormer22}, and SGFormer \cite{sgformer23}. Note that in our proposed CoBFormer, the GCN module (CoB-G) and BGA module (CoB-T) each predict the node label independently, so we report their performance simultaneously. Experimental implementation details are given in \cref{appendix:details}.

\textbf{Node Classification Results.} \cref{main-table}  reports the experimental results on node classification. We select Micro-F1 and Macro-F1 as metrics to conduct a comprehensive performance comparison. We can observe that: (1) Both GCN and BGA modules of CoBFormer outperform all baselines in homophilic graphs by a substantial margin, demonstrating the effectiveness of CoBFormer. (2) In heterophilic graphs, the performance of our BGA module is comparable to, even surpasses, the best baseline, SGFormer. This indicates that our BGA module can successfully capture global information.
(3) Traditional Graph Transformers exhibit superior performance on heterophilic graphs when compared with GCN and GAT. However, their advantage in homophilic graphs is relatively limited. This suggests that local information plays a more crucial role in homophilic graphs, whereas global information significantly enhances model performance in heterophilic graphs. These results are consistent with our analysis in \cref{over-globalizing}. We further conduct a significance test between our method and SGFormer, the best Transformer-based baseline. The results are provided in \cref{appendix:significance} and they indicate that our method not only significantly outperforms SGFormer in homophilic graphs but also achieves substantial improvements in heterophilic graphs.

\begin{table}[ht]
\vskip -0.1in
\caption{Test accuracy and GPU memory of various CoBFormer variants. `V-A' denotes the vanilla global attention. `B-A' represents the BGA module. `C-T' indicates whether collaborative training is applied.}
\label{ablation-table}
\vskip -0.15in
\begin{center}
\begin{small}
\resizebox{\columnwidth}{!}{
\begin{tabular}{l|ccc|cc|c}
\toprule
Dataset & V-A & B-A & C-T & CoB-G & CoB-T & MEM \\
\midrule
\multirow{4}{*}{Cora} & $\surd$  & $\times$  &$\times$   & 81.44 & 54.86 & 0.85G \\
& $\surd$  & $\times$  &$\surd$   & 83.78 & 83.82 & 0.85G\\
& $\times$  & $\surd$  &$\times$   & 81.44 & 68.72 & 0.38G \\
& $\times$  & $\surd$  &$\surd$   & 84.96 & 85.28 & 0.38G \\
\midrule
\multirow{4}{*}{PubMed} & $\surd$  & $\times$  &$\times$   & 79.26 & 71.22 & 8.42G \\
& $\surd$  & $\times$  &$\surd$   & 80.38 & 80.36 & 8.42G \\
& $\times$  & $\surd$  &$\times$   & 79.26 & 74.52 & 0.50G \\
& $\times$  & $\surd$  &$\surd$   & 80.52 & 81.42 & 0.50G \\
\midrule
\multirow{4}{*}{Deezer} & $\surd$  & $\times$  &$\times$   & 62.07 & 66.49 & 20.23G \\
& $\surd$  & $\times$  &$\surd$   & 63.67 & 66.86 & 20.23G \\
& $\times$  & $\surd$  &$\times$   & 62.07 & 66.56 & 3.97G \\
& $\times$  & $\surd$  &$\surd$   & 63.76 & 66.96 & 3.97G \\
\bottomrule
\end{tabular}}
\end{small}
\end{center}
\vskip -0.15in
\end{table}
\section{Ablation Studies \& Analysis}
\textbf{Ablation Study.} We carry out ablation studies on Cora, PubMed, and Deezer to evaluate the two fundamental components of our CoBFormer: the BGA module and the collaborative training approach, where the results are shown in \cref{ablation-table}. Key observations include: (1) The accuracy of our BGA module consistently outperforms vanilla global attention on all datasets, irrespective of the use of collaborative training, demonstrating the effectiveness of our BGA module. (2) Collaborative training leads to significant accuracy improvement in both the GCN and BGA modules, indicating that it enhances the model's generalization ability by encouraging mutual learning. (3) The BGA module significantly reduces GPU memory, addressing scalability concerns. Specifically, GPU memory usage is largely reduced by 94\% for PubMed and 80\% for Deezer.

\begin{figure}[htb]
\vskip -0.1in
\begin{center}
\centerline{\includegraphics[width=0.95\columnwidth]{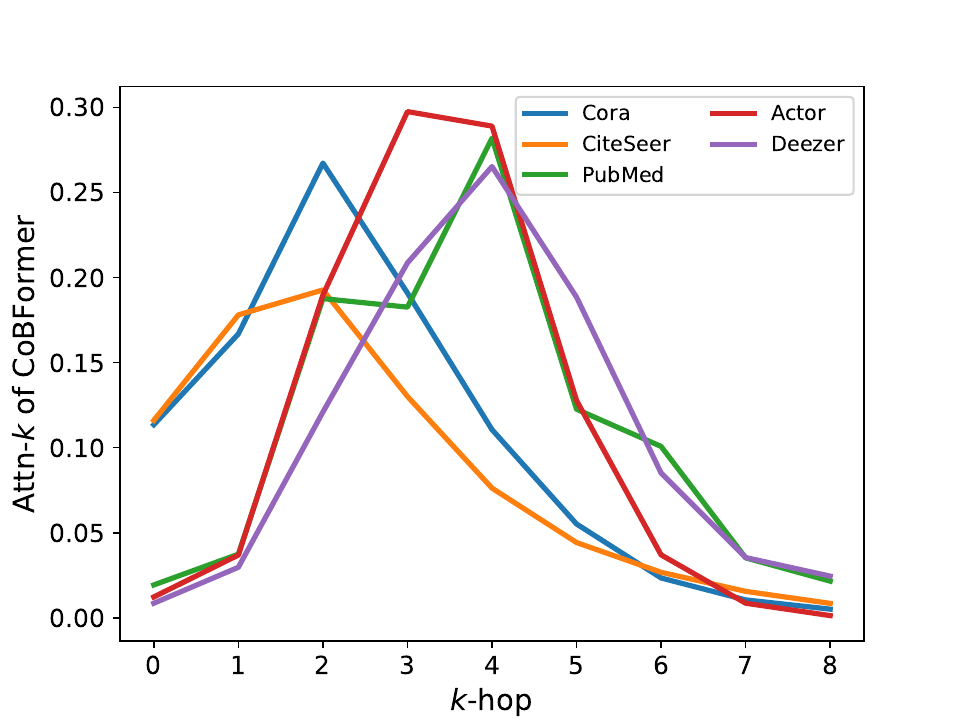}}
\caption{The Attn-$k$ of CoBFormer.}
\label{fig:attn_k_CoBi}
\end{center}
\vskip -0.2in
\end{figure}
\textbf{Over-Globalizing Problem.} To demonstrate our CoBFormer's ability to alleviate the over-globalizing problem, we visualize the Attn-$k$ of our CoBFormer in \cref{fig:attn_k_CoBi}. Compared with \cref{fig:attn_k_vt,fig:attn_k_node}, CoBFormer allocates more attention scores in the local region than Vanilla Transformer and NodeFormer, indicating that our BGA module can effectively alleviate the over-globalizing problem by decoupling the intra-cluster information and inter-cluster information.

\begin{figure}[htb]
\begin{center}
\centerline{\includegraphics[width=\columnwidth]{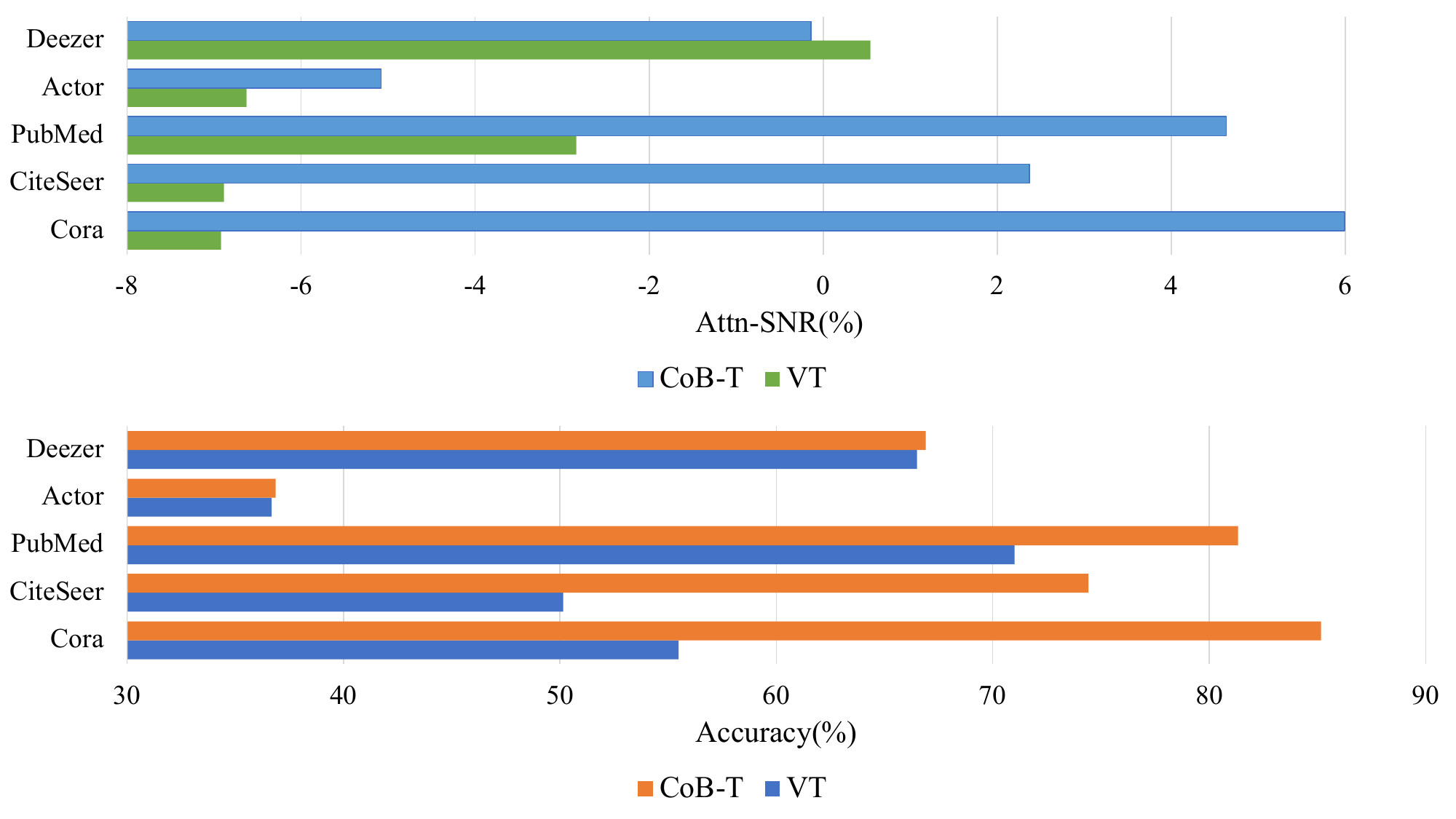}}
\caption{The Attn-SNR and Accuracy of VT and our CoBFormer.}
\label{fig:analysis_noise}
\end{center}
\vskip -0.2in
\end{figure}

We further calculate the Attn-SNR and test accuracy to show our model's capabilities to distinguish useful nodes and extract valuable information, where the results are shown in \cref{fig:analysis_noise}. Obviously, the CoB-T significantly improves the Attn-SNR and substantially boosts performance on the Cora, CiteSeer, and PubMed. It underscores that our CoBFormer can effectively mitigate the over-globalizing problem. For Actor and Deezer, our CoB-T achieves performance comparable to that of VT, implying that our CoBFormer can effectively capture global information.

\begin{figure}[htb]
\begin{center}
\centerline{\includegraphics[width=\columnwidth]{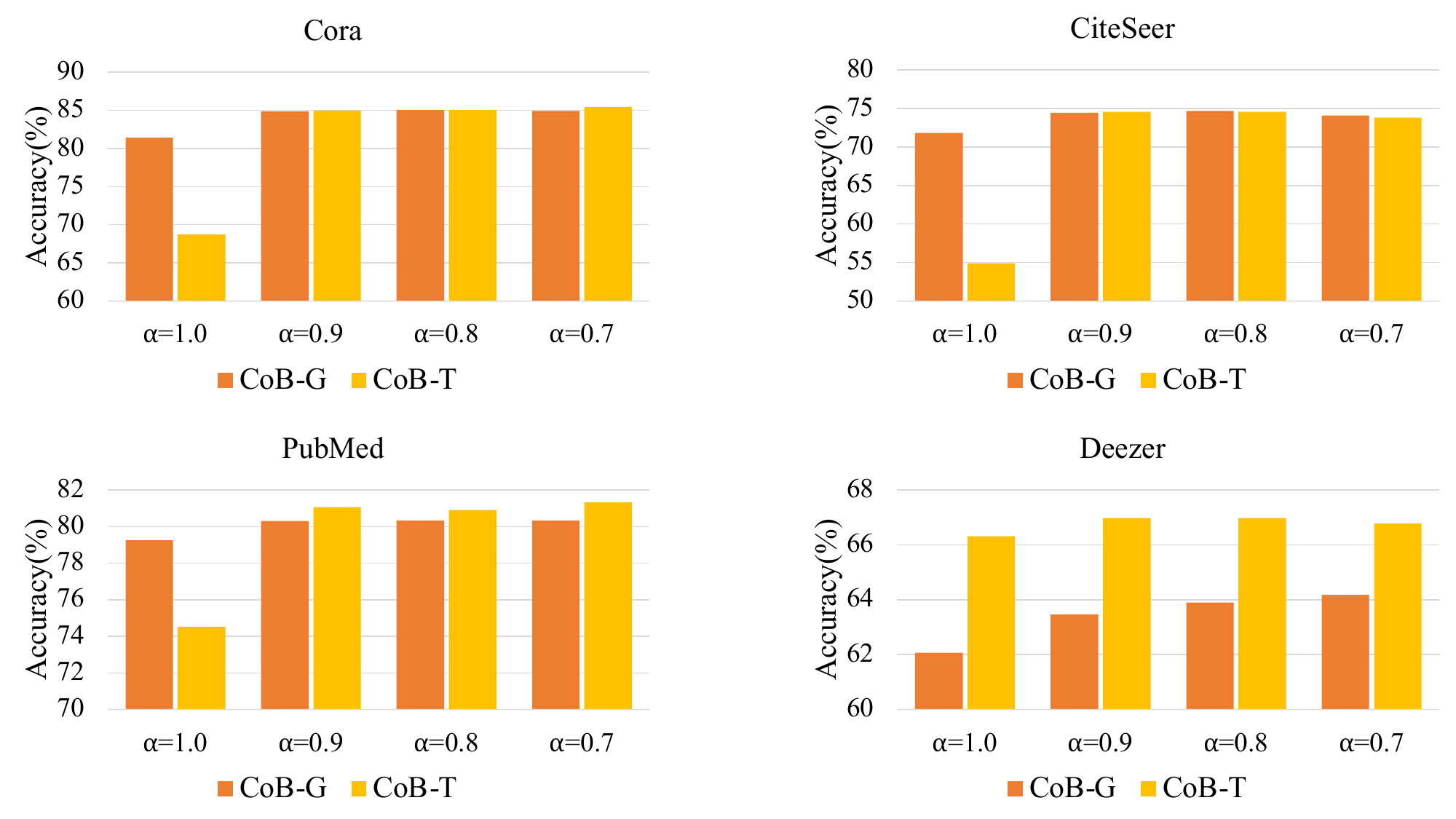}}
\caption{The average test accuracy of CoBFormer for different $\alpha$.}
\label{fig:ablation_alpha}
\end{center}
\vskip -0.2in
\end{figure}
\begin{figure}[htb]
\begin{center}
\centerline{\includegraphics[width=\columnwidth]{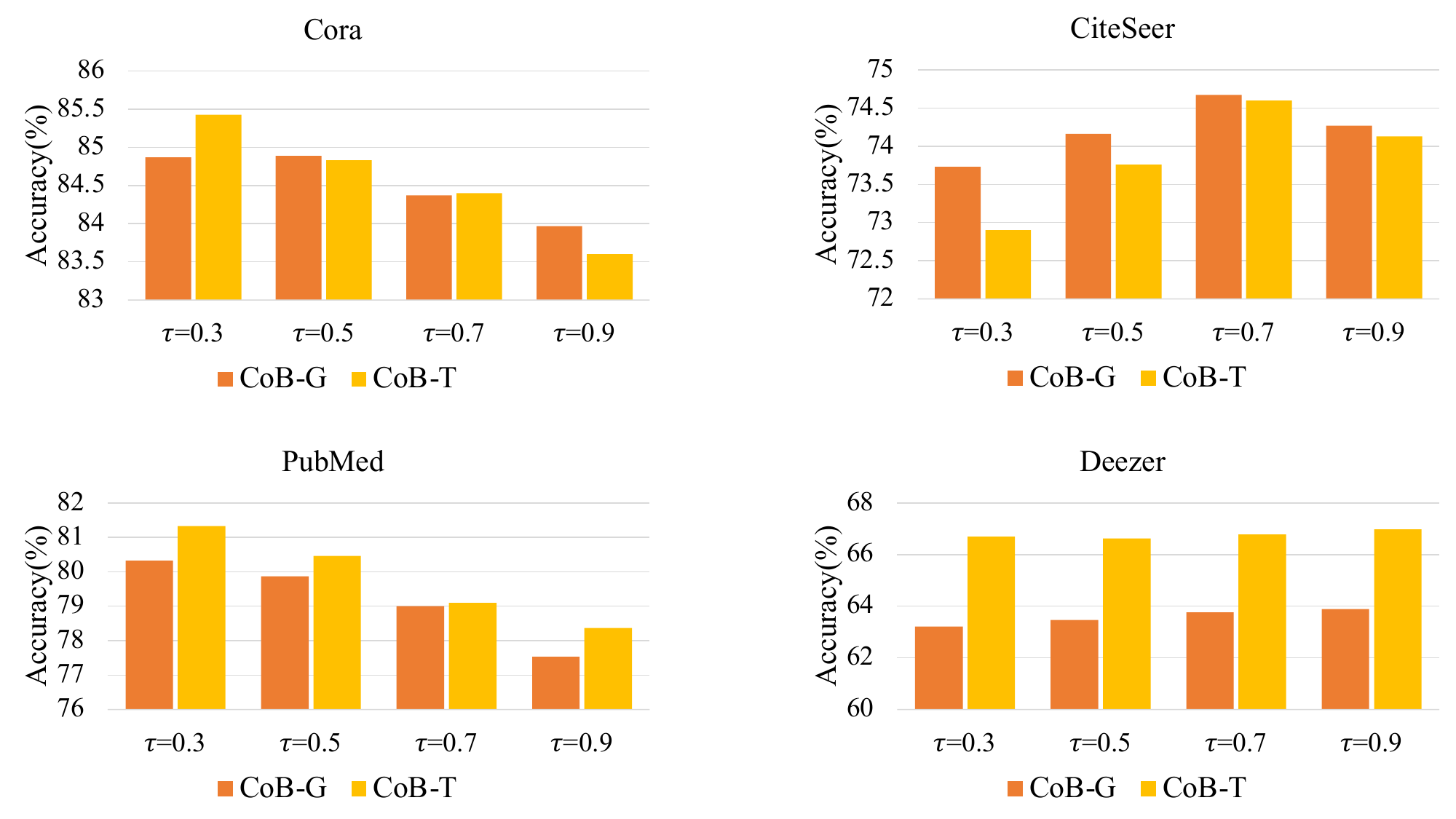}}
\caption{The average test accuracy of CoBFormer for different $\tau$.}
\label{fig:ablation_tau}
\end{center}
\vskip -0.35in
\end{figure}
\begin{figure}[htb]
\vskip -0.15in
\begin{center}
\centerline{\includegraphics[width=\columnwidth]{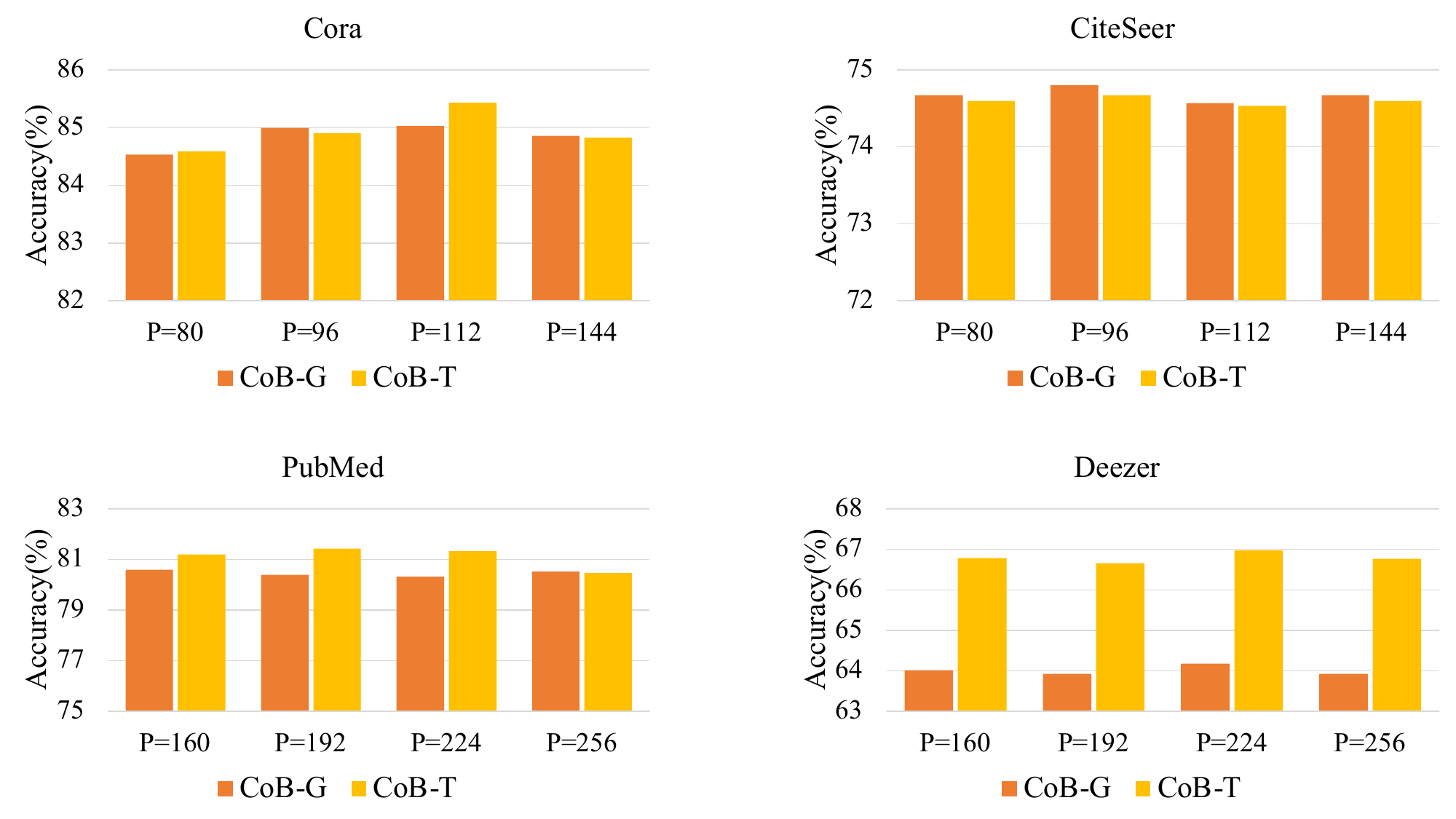}}
\caption{The average test accuracy of CoBFormer for different P.}
\label{fig:ablation_cluster}
\end{center}
\vskip -0.3in
\end{figure}
\textbf{Parameter Study.} We analyze the key parameters: the collaborative learning strength coefficient $\alpha$, the temperature coefficient $\tau$, and the number of clusters P. First, we vary $\alpha$ in $\{1.0, 0.9, 0.8, 0.7\}$ and report the best performance in \cref{fig:ablation_alpha}. Our results show that our model achieves notable performance improvements for all values except when $\alpha = 1$, in which case the collaborative training method is not employed. Furthermore, it exhibits consistent performance across the other $\alpha$ values, underscoring the effectiveness and robustness of our collaborative training approach. Next, we fix the optimal $\alpha$ and report the best performance for various $\tau$ values in $\{0.9, 0.7, 0.5, 0.3\}$. \cref{fig:ablation_tau} suggests that the choice of $\tau$ significantly impacts performance, emphasizing the importance of selecting an appropriate $\tau$ for optimal results. Finally, we fix the best $\alpha$ and $\tau$ values and select different numbers of clusters to report the classification accuracy in \cref{fig:ablation_cluster}. The results indicate that on Deezer, performance changes minimally with different cluster counts, whereas for other datasets, optimal cluster selection is vital for peak performance. Nevertheless, the model typically surpasses others even without ideal cluster configurations, showcasing its robustness across various cluster counts.


\section{Related Work}
\textbf{Graph Neural Networks.} Representative GNNs, such as GCN \cite{GCN17} and GAT \cite{GAT18}, leverage a message-passing mechanism to recursively aggregate neighbor information. However, due to over-smoothing \cite{over-smoothing18, over-smoothing19, over-smoothing20} and over-squashing \cite{over-squashing21,over-squashing22} problems, GNNs typically cannot stack multiple layers to capture information from distant nodes. Moreover, early designs of GNNs largely rely on the homophily assumption \cite{homo_assumpt01} that nodes of the same type are more likely to be connected. Although some GNNs are designed for heterophilic graphs \cite{Geom-GCN20,linkx21,GPR21,FAGCN21}, they still suffer from the same issues of over-smoothing and over-squashing, resulting in a limited receptive field.

\textbf{Graph Transformers.} Transformers \cite{transformer17}, benefiting from their global attention, naturally construct a fully connected graph with learnable edge weights, offering a novel approach to address the issues of over-smoothing and over-squashing in GNNs. Extensive works have achieved remarkable success in graph-level tasks \cite{graphormer21,san21,graphtrans21,graphgps22}. This success is primarily attributed to their global perception capability, which is vital for graph-level tasks. Influenced by the success in graph-level tasks, researchers are now exploring the integration of the global attention mechanism into node-level tasks \cite{coarformer22,ans-gt22,hsgt23,gapformer23,nodeformer22, GOAT23, sgformer23}. These approaches aim to broaden the receptive field in large-scale graphs and amplify the expressive potential of the models.

\section{Conclusion}
In this paper, we discover the over-globalizing problem in Graph Transformers by presenting the theoretical insights and empirical results. We then propose CoBFormer, a bi-level global graph transformer with collaborative training, aiming at alleviating the over-globalizing problem and improving the generalization ability. Extensive experiments verify the effectiveness of CoBFormer.

\section*{Acknowledgements}
This work is supported in part by the National Natural Science Foundation of China (No. U20B2045, 62322203, 62172052, 62192784, U22B2038).

\section*{Impact Statement}
This paper presents work whose goal is to advance the field of Graph Machine Learning and will promote the application of graph machine learning in large-scale graph data mining, such as enhancing social network recommendations and reducing traffic network congestion. There are many potential societal consequences of our work, none which we feel must be specifically highlighted here.


\bibliography{icml2024}

\begin{thebibliography}{44}
\providecommand{\natexlab}[1]{#1}
\providecommand{\url}[1]{\texttt{#1}}
\expandafter\ifx\csname urlstyle\endcsname\relax
  \providecommand{\doi}[1]{doi: #1}\else
  \providecommand{\doi}{doi: \begingroup \urlstyle{rm}\Url}\fi

\bibitem[Ba et~al.(2016)Ba, Kiros, and Hinton]{layernorm}
Ba, J.~L., Kiros, J.~R., and Hinton, G.~E.
\newblock Layer normalization.
\newblock \emph{arXiv preprint arXiv:1607.06450}, 2016.

\bibitem[Bo et~al.(2021)Bo, Wang, Shi, and Shen]{FAGCN21}
Bo, D., Wang, X., Shi, C., and Shen, H.
\newblock Beyond low-frequency information in graph convolutional networks.
\newblock In \emph{Proceedings of the AAAI Conference on Artificial Intelligence}, volume~35, pp.\  3950--3957, 2021.

\bibitem[Chen et~al.(2022)Chen, Gao, Li, and He]{NAGphormer22}
Chen, J., Gao, K., Li, G., and He, K.
\newblock Nagphormer: A tokenized graph transformer for node classification in large graphs.
\newblock In \emph{The Eleventh International Conference on Learning Representations}, 2022.

\bibitem[Chien et~al.(2021)Chien, Peng, Li, and Milenkovic]{GPR21}
Chien, E., Peng, J., Li, P., and Milenkovic, O.
\newblock Adaptive universal generalized pagerank graph neural network.
\newblock In \emph{International Conference on Learning Representations}, 2021.

\bibitem[Choromanski et~al.(2020)Choromanski, Likhosherstov, Dohan, Song, Gane, Sarlos, Hawkins, Davis, Mohiuddin, Kaiser, et~al.]{performer20}
Choromanski, K., Likhosherstov, V., Dohan, D., Song, X., Gane, A., Sarlos, T., Hawkins, P., Davis, J., Mohiuddin, A., Kaiser, L., et~al.
\newblock Rethinking attention with performers.
\newblock \emph{arXiv preprint arXiv:2009.14794}, 2020.

\bibitem[Deac et~al.(2022)Deac, Lackenby, and Veli{\v{c}}kovi{\'c}]{over-squashing22}
Deac, A., Lackenby, M., and Veli{\v{c}}kovi{\'c}, P.
\newblock Expander graph propagation.
\newblock In \emph{Learning on Graphs Conference}, pp.\  38--1. PMLR, 2022.

\bibitem[Devlin et~al.(2018)Devlin, Chang, Lee, and Toutanova]{bert18}
Devlin, J., Chang, M.-W., Lee, K., and Toutanova, K.
\newblock Bert: Pre-training of deep bidirectional transformers for language understanding.
\newblock \emph{arXiv preprint arXiv:1810.04805}, 2018.

\bibitem[Dosovitskiy et~al.(2021)Dosovitskiy, Beyer, Kolesnikov, Weissenborn, Zhai, Unterthiner, Dehghani, Minderer, Heigold, Gelly, Uszkoreit, and Houlsby]{vit21}
Dosovitskiy, A., Beyer, L., Kolesnikov, A., Weissenborn, D., Zhai, X., Unterthiner, T., Dehghani, M., Minderer, M., Heigold, G., Gelly, S., Uszkoreit, J., and Houlsby, N.
\newblock An image is worth 16x16 words: Transformers for image recognition at scale.
\newblock In \emph{International Conference on Learning Representations}, 2021.

\bibitem[Fey \& Lenssen(2019)Fey and Lenssen]{pyg19}
Fey, M. and Lenssen, J.~E.
\newblock Fast graph representation learning with pytorch geometric.
\newblock \emph{arXiv preprint arXiv:1903.02428}, 2019.

\bibitem[Hamilton et~al.(2017)Hamilton, Ying, and Leskovec]{graphsage17}
Hamilton, W., Ying, Z., and Leskovec, J.
\newblock Inductive representation learning on large graphs.
\newblock \emph{Advances in neural information processing systems}, 30, 2017.

\bibitem[He et~al.(2016)He, Zhang, Ren, and Sun]{residual16}
He, K., Zhang, X., Ren, S., and Sun, J.
\newblock Deep residual learning for image recognition.
\newblock In \emph{Proceedings of the IEEE conference on computer vision and pattern recognition}, pp.\  770--778, 2016.

\bibitem[Hinton et~al.(2015)Hinton, Vinyals, and Dean]{kd15}
Hinton, G., Vinyals, O., and Dean, J.
\newblock Distilling the knowledge in a neural network.
\newblock \emph{arXiv preprint arXiv:1503.02531}, 2015.

\bibitem[Hu et~al.(2020)Hu, Fey, Zitnik, Dong, Ren, Liu, Catasta, and Leskovec]{ogb20}
Hu, W., Fey, M., Zitnik, M., Dong, Y., Ren, H., Liu, B., Catasta, M., and Leskovec, J.
\newblock Open graph benchmark: Datasets for machine learning on graphs.
\newblock \emph{Advances in neural information processing systems}, 33:\penalty0 22118--22133, 2020.

\bibitem[Kalofolias(2016)]{graphdenoise16}
Kalofolias, V.
\newblock How to learn a graph from smooth signals.
\newblock In \emph{Artificial intelligence and statistics}, pp.\  920--929. PMLR, 2016.

\bibitem[Karypis \& Kumar(1998)Karypis and Kumar]{metis98}
Karypis, G. and Kumar, V.
\newblock A fast and high quality multilevel scheme for partitioning irregular graphs.
\newblock \emph{SIAM Journal on scientific Computing}, 20\penalty0 (1):\penalty0 359--392, 1998.

\bibitem[Kipf \& Welling(2017)Kipf and Welling]{GCN17}
Kipf, T.~N. and Welling, M.
\newblock Semi-supervised classification with graph convolutional networks.
\newblock In \emph{International Conference on Learning Representations}, 2017.

\bibitem[Kong et~al.(2023)Kong, Chen, Kirchenbauer, Ni, Bruss, and Goldstein]{GOAT23}
Kong, K., Chen, J., Kirchenbauer, J., Ni, R., Bruss, C.~B., and Goldstein, T.
\newblock Goat: A global transformer on large-scale graphs.
\newblock In \emph{International Conference on Machine Learning}, pp.\  17375--17390. PMLR, 2023.

\bibitem[Kreuzer et~al.(2021)Kreuzer, Beaini, Hamilton, L{\'e}tourneau, and Tossou]{san21}
Kreuzer, D., Beaini, D., Hamilton, W., L{\'e}tourneau, V., and Tossou, P.
\newblock Rethinking graph transformers with spectral attention.
\newblock \emph{Advances in Neural Information Processing Systems}, 34:\penalty0 21618--21629, 2021.

\bibitem[Kuang et~al.(2022)Kuang, WANG, Li, Wei, and Ding]{coarformer22}
Kuang, W., WANG, Z., Li, Y., Wei, Z., and Ding, B.
\newblock Coarformer: Transformer for large graph via graph coarsening, 2022.

\bibitem[Li et~al.(2018)Li, Han, and Wu]{over-smoothing18}
Li, Q., Han, Z., and Wu, X.-M.
\newblock Deeper insights into graph convolutional networks for semi-supervised learning.
\newblock In \emph{Proceedings of the AAAI conference on artificial intelligence}, volume~32, 2018.

\bibitem[Lim et~al.(2021{\natexlab{a}})Lim, Hohne, Li, Huang, Gupta, Bhalerao, and Lim]{linkx21}
Lim, D., Hohne, F.~M., Li, X., Huang, S.~L., Gupta, V., Bhalerao, O.~P., and Lim, S.-N.
\newblock Large scale learning on non-homophilous graphs: New benchmarks and strong simple methods.
\newblock In Beygelzimer, A., Dauphin, Y., Liang, P., and Vaughan, J.~W. (eds.), \emph{Advances in Neural Information Processing Systems}, 2021{\natexlab{a}}.

\bibitem[Lim et~al.(2021{\natexlab{b}})Lim, Li, Hohne, and Lim]{heterophilic21}
Lim, D., Li, X., Hohne, F., and Lim, S.-N.
\newblock New benchmarks for learning on non-homophilous graphs.
\newblock \emph{arXiv preprint arXiv:2104.01404}, 2021{\natexlab{b}}.

\bibitem[Liu et~al.(2023{\natexlab{a}})Liu, Zhan, Ma, Ding, Tao, Wu, and Hu]{gapformer23}
Liu, C., Zhan, Y., Ma, X., Ding, L., Tao, D., Wu, J., and Hu, W.
\newblock Gapformer: Graph transformer with graph pooling for node classification.
\newblock In Elkind, E. (ed.), \emph{Proceedings of the Thirty-Second International Joint Conference on Artificial Intelligence, {IJCAI-23}}, pp.\  2196--2205. International Joint Conferences on Artificial Intelligence Organization, 8 2023{\natexlab{a}}.
\newblock Main Track.

\bibitem[Liu et~al.(2023{\natexlab{b}})Liu, Yang, Zhao, Han, Zhang, Wu, Zhou, Huang, Wang, and Shi]{gammagl}
Liu, Y., Yang, C., Zhao, T., Han, H., Zhang, S., Wu, J., Zhou, G., Huang, H., Wang, H., and Shi, C.
\newblock Gammagl: A multi-backend library for graph neural networks.
\newblock In \emph{Proceedings of the 46th International ACM SIGIR Conference on Research and Development in Information Retrieval}, pp.\  2861--2870, 2023{\natexlab{b}}.

\bibitem[McPherson et~al.(2001)McPherson, Smith-Lovin, and Cook]{homo_assumpt01}
McPherson, M., Smith-Lovin, L., and Cook, J.~M.
\newblock Birds of a feather: Homophily in social networks.
\newblock \emph{Annual review of sociology}, 27:\penalty0 415--444, 2001.

\bibitem[Nt \& Maehara(2019)Nt and Maehara]{over-smoothing19}
Nt, H. and Maehara, T.
\newblock Revisiting graph neural networks: All we have is low-pass filters.
\newblock \emph{arXiv preprint arXiv:1905.09550}, 2019.

\bibitem[Oono \& Suzuki(2020)Oono and Suzuki]{over-smoothing20}
Oono, K. and Suzuki, T.
\newblock Graph neural networks exponentially lose expressive power for node classification.
\newblock In \emph{International Conference on Learning Representations}, 2020.

\bibitem[Pei et~al.(2020)Pei, Wei, Chang, Lei, and Yang]{Geom-GCN20}
Pei, H., Wei, B., Chang, K. C.-C., Lei, Y., and Yang, B.
\newblock Geom-gcn: Geometric graph convolutional networks.
\newblock In \emph{International Conference on Learning Representations}, 2020.

\bibitem[Ramp{\'a}{\v{s}}ek et~al.(2022)Ramp{\'a}{\v{s}}ek, Galkin, Dwivedi, Luu, Wolf, and Beaini]{graphgps22}
Ramp{\'a}{\v{s}}ek, L., Galkin, M., Dwivedi, V.~P., Luu, A.~T., Wolf, G., and Beaini, D.
\newblock Recipe for a general, powerful, scalable graph transformer.
\newblock \emph{Advances in Neural Information Processing Systems}, 35:\penalty0 14501--14515, 2022.

\bibitem[Shuman et~al.(2013)Shuman, Narang, Frossard, Ortega, and Vandergheynst]{graphdenoise13}
Shuman, D.~I., Narang, S.~K., Frossard, P., Ortega, A., and Vandergheynst, P.
\newblock The emerging field of signal processing on graphs: Extending high-dimensional data analysis to networks and other irregular domains.
\newblock \emph{IEEE signal processing magazine}, 30\penalty0 (3):\penalty0 83--98, 2013.

\bibitem[Topping et~al.(2021)Topping, Di~Giovanni, Chamberlain, Dong, and Bronstein]{over-squashing21}
Topping, J., Di~Giovanni, F., Chamberlain, B.~P., Dong, X., and Bronstein, M.~M.
\newblock Understanding over-squashing and bottlenecks on graphs via curvature.
\newblock \emph{arXiv preprint arXiv:2111.14522}, 2021.

\bibitem[Vaswani et~al.(2017)Vaswani, Shazeer, Parmar, Uszkoreit, Jones, Gomez, Kaiser, and Polosukhin]{transformer17}
Vaswani, A., Shazeer, N., Parmar, N., Uszkoreit, J., Jones, L., Gomez, A.~N., Kaiser, {\L}., and Polosukhin, I.
\newblock Attention is all you need.
\newblock \emph{Advances in neural information processing systems}, 30, 2017.

\bibitem[Veličković et~al.(2018)Veličković, Cucurull, Casanova, Romero, Liò, and Bengio]{GAT18}
Veličković, P., Cucurull, G., Casanova, A., Romero, A., Liò, P., and Bengio, Y.
\newblock Graph attention networks.
\newblock In \emph{International Conference on Learning Representations}, 2018.

\bibitem[Wang et~al.(2020)Wang, Li, Khabsa, Fang, and Ma]{linformer20}
Wang, S., Li, B.~Z., Khabsa, M., Fang, H., and Ma, H.
\newblock Linformer: Self-attention with linear complexity.
\newblock \emph{arXiv preprint arXiv:2006.04768}, 2020.

\bibitem[Wu et~al.(2022)Wu, Zhao, Li, Wipf, and Yan]{nodeformer22}
Wu, Q., Zhao, W., Li, Z., Wipf, D.~P., and Yan, J.
\newblock Nodeformer: A scalable graph structure learning transformer for node classification.
\newblock \emph{Advances in Neural Information Processing Systems}, 35:\penalty0 27387--27401, 2022.

\bibitem[Wu et~al.(2023)Wu, Zhao, Yang, Zhang, Nie, Jiang, Bian, and Yan]{sgformer23}
Wu, Q., Zhao, W., Yang, C., Zhang, H., Nie, F., Jiang, H., Bian, Y., and Yan, J.
\newblock Simplifying and empowering transformers for large-graph representations.
\newblock In \emph{Thirty-seventh Conference on Neural Information Processing Systems}, 2023.

\bibitem[Wu et~al.(2021)Wu, Jain, Wright, Mirhoseini, Gonzalez, and Stoica]{graphtrans21}
Wu, Z., Jain, P., Wright, M., Mirhoseini, A., Gonzalez, J.~E., and Stoica, I.
\newblock Representing long-range context for graph neural networks with global attention.
\newblock \emph{Advances in Neural Information Processing Systems}, 34:\penalty0 13266--13279, 2021.

\bibitem[Yang et~al.(2017)Yang, Sun, Zhao, Liu, and Chang]{yang2017neural}
Yang, C., Sun, M., Zhao, W.~X., Liu, Z., and Chang, E.~Y.
\newblock A neural network approach to jointly modeling social networks and mobile trajectories.
\newblock \emph{ACM Transactions on Information Systems (TOIS)}, 35\penalty0 (4):\penalty0 1--28, 2017.

\bibitem[Yang et~al.(2016)Yang, Cohen, and Salakhudinov]{citenetwork16}
Yang, Z., Cohen, W., and Salakhudinov, R.
\newblock Revisiting semi-supervised learning with graph embeddings.
\newblock In \emph{International conference on machine learning}, pp.\  40--48. PMLR, 2016.

\bibitem[Ying et~al.(2021)Ying, Cai, Luo, Zheng, Ke, He, Shen, and Liu]{graphormer21}
Ying, C., Cai, T., Luo, S., Zheng, S., Ke, G., He, D., Shen, Y., and Liu, T.-Y.
\newblock Do transformers really perform badly for graph representation?
\newblock \emph{Advances in Neural Information Processing Systems}, 34:\penalty0 28877--28888, 2021.

\bibitem[Yu et~al.(2023)Yu, Liu, Fang, and Zhang]{yu2023learning}
Yu, X., Liu, Z., Fang, Y., and Zhang, X.
\newblock Learning to count isomorphisms with graph neural networks.
\newblock In \emph{Proceedings of the AAAI Conference on Artificial Intelligence}, volume~37, pp.\  4845--4853, 2023.

\bibitem[Zhang et~al.(2022)Zhang, Liu, Hu, and Lee]{ans-gt22}
Zhang, Z., Liu, Q., Hu, Q., and Lee, C.-K.
\newblock Hierarchical graph transformer with adaptive node sampling.
\newblock \emph{Advances in Neural Information Processing Systems}, 35:\penalty0 21171--21183, 2022.

\bibitem[Zhao et~al.(2021)Zhao, Li, Wen, Wang, Liu, Sun, Xie, and Ye]{gophormer21}
Zhao, J., Li, C., Wen, Q., Wang, Y., Liu, Y., Sun, H., Xie, X., and Ye, Y.
\newblock Gophormer: Ego-graph transformer for node classification.
\newblock \emph{arXiv preprint arXiv:2110.13094}, 2021.

\bibitem[Zhu et~al.(2023)Zhu, Wen, Song, Ma, and Wang]{hsgt23}
Zhu, W., Wen, T., Song, G., Ma, X., and Wang, L.
\newblock Hierarchical transformer for scalable graph learning.
\newblock In Elkind, E. (ed.), \emph{Proceedings of the Thirty-Second International Joint Conference on Artificial Intelligence, {IJCAI-23}}, pp.\  4702--4710. International Joint Conferences on Artificial Intelligence Organization, 8 2023.
\newblock Main Track.

\end{thebibliography}
\bibliographystyle{icml2024}

\newpage
\appendix
\onecolumn
\section{Proofs}
\label{appendix:proof}
\subsection{Proof of \cref{thm:denoise2attn}}
\label{appendix:proof1}
\textit{Proof:} To prove \cref{thm:denoise2attn}, we assume the existence of a linear classifier, parameterized by $\mathbf{W}_C$, which satisfies the condition $\mathbf{ZW}_C=\mathbf{Y}$. Consequently, this leads to $\mathbf{Z}=\mathbf{Y}\mathbf{W}_C^{-1}$.
\begin{equation}
    \begin{aligned}
    \| \mathbf{Z} - \mathbf{\hat AZ} \|_F 
    &\le \sum_{u \in \mathcal{V}} \| \mathbf{z}_u - \sum_{v \in \mathcal{V}} \alpha_{uv}\mathbf{z}_v  \|_2  
    	&&\cdots \sqrt{a+b+c}\le\sqrt{a} + \sqrt{b} + \sqrt{c} \\
    &= \sum_{u \in \mathcal{V}} \|\sum_{v \in \mathcal{V}} \alpha_{uv}(\mathbf{z}_u-\mathbf{z}_v)  \|_2  
    	&&\cdots \sum_{v \in \mathcal{V}} \alpha_{uv}=1 \\
    &\le \sum_{u,v \in \mathcal{V}} \alpha_{u,v} \| \mathbf{z}_u-\mathbf{z}_v \|_2 
    	&&\cdots \text{The Triangle Inequality} \\
    &\le L\sum_{u,v \in \mathcal{V}} \alpha_{u,v} \| \mathbf{y}_u-\mathbf{y}_v \|_2 
    	&&\cdots \text{Lipschitz Continuous} \\
    &= L\sum_{u \in \mathcal{V}}\sum_{v \in \mathcal{V},y_v = y_u} \alpha_{u,v} \| \mathbf{y}_u-\mathbf{y}_v \|_2
     + L\sum_{u \in \mathcal{V}}\sum_{v \in \mathcal{V},y_v \ne y_u} \alpha_{u,v} \| \mathbf{y}_u-\mathbf{y}_v \|_2 \\
    &= \sqrt{2}L\sum_{u \in \mathcal{V}}\sum_{v \in \mathcal{V},y_v \ne y_u} \alpha_{u,v}.
    \end{aligned}
\end{equation}
$L$ is a Lipschitz constant. For a given node $u$ and a well trained Graph Transformer, let $\eta_u=\mathbb{E}_{v \in \mathcal{V},\mathbf{y}_u = \mathbf{y}_v}\exp(\frac{\mathbf{q}_u\mathbf{k}_v^T}{\sqrt{d}})$, $\gamma_u=\mathbb{E}_{v \in \mathcal{V},\mathbf{y}_u \neq \mathbf{y}_v}\exp(\frac{\mathbf{q}_u\mathbf{k}_v^T}{\sqrt{d}})$. Let $\mathcal{C}_u$ represent the proportion of nodes that are reachable by node $u$ and belong to the same class. Then the theory  can be rewritten as:
\begin{equation}
    \begin{aligned}
    \| \mathbf{Z} - \mathbf{\hat AZ} \|_F &\le \sqrt{2}L\sum_{u \in \mathcal{V}}\sum_{v \in \mathcal{V},y_v \ne y_u} \alpha_{uv} \\
    &= \sqrt{2}L\sum_{u \in \mathcal{V}} \frac{\sum_{v \in \mathcal{V},y_v \ne y_u} \gamma_u}{\sum_{v \in \mathcal{V},y_v = y_u} \eta_u +\sum_{v \in \mathcal{V},y_v \ne y_u}\gamma_u}\\
    &= \sqrt{2}L\sum_{u \in \mathcal{V}}\frac{N(1-\mathcal{C}_u)\gamma_u}{N\mathcal{C}_u\eta_u+N(1-\mathcal{C}_u)\gamma_u} \\
    &= \sqrt{2}L\sum_{u \in \mathcal{V}}\frac{1}{1+\frac{\mathcal{C}_u}{1-\mathcal{C}_u}\frac{\eta_u}{\gamma_u}}. \\
    \end{aligned}
\end{equation}

\subsection{Proof of \cref{thm:same-label-rate}}
\label{appendix:proof2}
\textit{Proof:} We outline the proof of \cref{thm:same-label-rate} through the following several steps:

\begin{itemize}
    \item When $k=0$ or $k=1$, the equation is obviously true.
    \item When $k>1$, the derivation is as follows:
       \begin{equation}
           \begin{aligned}
           \mathcal{C}_u^k &= \mathcal{C}_u^{k-1}\rho+\frac{(1-\mathcal{C}_u^{k-1})(1-\rho)}{|\mathcal{Y}|-1} \\
           &= \frac{\mathcal{C}_u^{k-1}\rho(|\mathcal{Y}|-1)+(1-\mathcal{C}_u^{k-1})(1-\rho)}{|\mathcal{Y}|-1} \\
           &= \frac{1+|\mathcal{Y}|\rho\mathcal{C}_u^{k-1}-\rho-\mathcal{C}_u^{k-1}}{|\mathcal{Y}|-1}.
           \end{aligned}
       \end{equation}
    \item Then, we have:
        \begin{equation}
            \begin{aligned}
               \mathcal{C}_u^{k+1}-\mathcal{C}_u^{k} &= \frac{1+|\mathcal{Y}|\rho\mathcal{C}_u^k-\rho-\mathcal{C}_u^k}{|\mathcal{Y}|-1} - \mathcal{C}_u^{k} \\
               &= \frac{1+|\mathcal{Y}|\rho\mathcal{C}_u^k-\rho-\mathcal{C}_u^k|\mathcal{Y}|}{|\mathcal{Y}|-1} \\
               &= \frac{(1-\rho)(1-\mathcal{C}_u^k|\mathcal{Y}|)}{|\mathcal{Y}|-1}.
            \end{aligned}
        \end{equation}
        Notice that when $\mathcal{C}_u^k \ge \frac{1}{|\mathcal{Y}|}$, the right side of the equation becomes less than $0$, leading to $\mathcal{C}_u^k \ge \mathcal{C}_u^{k+1}$. Conversely, when $\mathcal{C}_u^k < \frac{1}{|\mathcal{Y}|}$, the right side is greater than $0$, resulting in $\mathcal{C}_u^k < \mathcal{C}_u^{k+1}$.
    \item Now, let us study the relation between $\mathcal{C}_u^k$ and $\frac{1}{|\mathcal{Y}|}$.
        \begin{equation}
            \begin{aligned}
                \mathcal{C}_u^k - \frac{1}{|\mathcal{Y}|} &= \frac{1+|\mathcal{Y}|\rho\mathcal{C}_u^{k-1}-\rho-\mathcal{C}_u^{k-1}}{|\mathcal{Y}|-1} - \frac{1}{|\mathcal{Y}|} \\
                &= \frac{(|\mathcal{Y}|\rho-1)(|\mathcal{Y}|\mathcal{C}_u^{k-1}-1)}{|\mathcal{Y}-1|\cdot|\mathcal{Y}|}.
            \end{aligned}
        \end{equation}
        When $\rho \ge \frac{1}{|\mathcal{Y}|}$ and $\mathcal{C}_u^{k-1} \ge \frac{1}{|\mathcal{Y}|}$, $\mathcal{C}_u^{k}$ will also be greater than $\frac{1}{|\mathcal{Y}|}$. This is always true when $\rho \ge \frac{1}{|\mathcal{Y}|}$. However, if $\rho < \frac{1}{|\mathcal{Y}|}$, then $\mathcal{C}_u^{k}$ will be greater than $\frac{1}{|\mathcal{Y}|}$ if $\mathcal{C}_u^{k-1} < \frac{1}{|\mathcal{Y}|}$, and less than $\frac{1}{|\mathcal{Y}|}$ if $\mathcal{C}_u^{k-1} > \frac{1}{|\mathcal{Y}|}$.
    \item Moreover, when we take into account both $\mathcal{C}_u^k$ and $\mathcal{C}_u^{k+2}$, we observe the following:
        \begin{equation}
            \begin{aligned}
               \mathcal{C}_u^{k+2}-\mathcal{C}_u^{k} &= \mathcal{C}_u^{k+2}-\mathcal{C}_u^{k+1} + \mathcal{C}_u^{k+1}-\mathcal{C}_u^{k} \\
               &= \frac{(1-\rho)(1-\mathcal{C}_u^{k+1}|\mathcal{Y}|)}{|\mathcal{Y}|-1} + \frac{(1-\rho)(1-\mathcal{C}_u^k|\mathcal{Y}|)}{|\mathcal{Y}|-1} \\
               &= \frac{(1-\rho)(2-(\mathcal{C}_u^k+\mathcal{C}_u^{k+2})|\mathcal{Y}|)}{|\mathcal{Y}|-1}.
            \end{aligned}
        \end{equation}
        Note that both $\mathcal{C}_u^k$ and $\mathcal{C}_u^{k+2}$ are either greater than $\frac{1}{|\mathcal{Y}|}$ or less than $\frac{1}{|\mathcal{Y}|}$. When they are greater than $\frac{1}{|\mathcal{Y}|}$, $\mathcal{C}_u^{k+2}$ will be less than $\mathcal{C}_u^k$. Conversely, when they are less than $\frac{1}{|\mathcal{Y}|}$, $\mathcal{C}_u^{k+2}$ will be greater than $\mathcal{C}_u^k$.
\end{itemize}
In summary, \cref{thm:same-label-rate} is proven through the aforementioned derivations.

\subsection{Proof of \cref{thm:co-train}}
\label{appendix:proof3}
\textit{Proof:}
We prove \cref{thm:co-train} using a technique of variational inference.
\begin{equation}
    \begin{aligned}
    \mathbb{E}_{P(\mathbf{L},\mathbf{U})}\log P_G(\mathbf{L},\mathbf{U})&=\int_{\mathbf{L},\mathbf{U}} P(\mathbf{L})P(\mathbf{U}|\mathbf{L})\log P_G(\mathbf{L},\mathbf{U}) \ d\mathbf{L} d\mathbf{U} \\
    &=\int_{\mathbf{L},\mathbf{U}} P(\mathbf{L})P(\mathbf{U}|\mathbf{L})\log \frac{P_G(\mathbf{L},\mathbf{U})P_T(\mathbf{U}|\mathbf{L})}{P_T(\mathbf{U}|\mathbf{L})} \ d\mathbf{L} d\mathbf{U} \\
    &\approx \int_{\mathbf{L},\mathbf{U}} P(\mathbf{L})P_T(\mathbf{U}|\mathbf{L})\log \frac{P_G(\mathbf{L},\mathbf{U})P(\mathbf{U}|\mathbf{L})}{P_T(\mathbf{U}|\mathbf{L}))} \ d\mathbf{L} d\mathbf{U} \\
    &= \mathbb{E}_{P(\mathbf{L})P_T(\mathbf{U}|\mathbf{L})}\log\frac{P_G(\mathbf{L})P_G(\mathbf{U}|\mathbf{L})P(\mathbf{U}|\mathbf{L})}{P_T(\mathbf{U}|\mathbf{L})} \\
    &= \mathbb{E}_{P(\mathbf{L})}\log P_G(\mathbf{L}) + \mathbb{E}_{P_T(\mathbf{U}|\mathbf{L})}\log P_G(\mathbf{U}|\mathbf{L}) + \mathbb{E}_{P_T(\mathbf{U}|\mathbf{L})}\log \frac{P(\mathbf{U}|\mathbf{L})}{P_T(\mathbf{U}|\mathbf{L})} \\
    &= \mathbb{E}_{P(\mathbf{L})}\log P_G(\mathbf{L}) + \mathbb{E}_{P_T(\mathbf{U}|\mathbf{L})}\log P_G(\mathbf{U}|\mathbf{L})-\text{KL}( P_T(\mathbf{U}|\mathbf{L}) \| P(\mathbf{U}|\mathbf{L})).
    \end{aligned}
\end{equation}
The proof of the latter follows by the same reasoning.

\subsection{Proof of \cref{prop:approximate-global}}
\label{appendix:proof4}
\textit{Proof:}
Through the inter-cluster Transformer and the concatenation operation, a node $u$ in cluster $p$ is enabled to access information from cluster $q$ by computing $\dot \alpha_{pq}\mathbf{v}'_q$. Here, $\mathbf{v}'_q$ is defined as $\mathbf{p}_p^k\mathbf{W}^k_{V'}$, where $\mathbf{p}_p^k$ is derived through a mean pooling operation, specifically $\mathbf{p}_p^k=\frac{\sum_{v \in \mathcal{V}_q}\mathbf{h}_v}{|\mathcal{V}_q|}$. Consequently, we can express the process as follows:
\begin{equation}
\label{eq:approx}
    \begin{aligned}
	&\dot \alpha_{pq}\mathbf{v}_q' \\
     =&\dot \alpha_{pq}\mathbf{p}_p^k\mathbf{W}^k_{V'} \\
     =&\dot \alpha_{pq}\frac{\sum_{v \in \mathcal{V}_q}\mathbf{h}_v}{|\mathcal{V}_q|}\mathbf{W}^k_{V'} \\
     =&\sum_{v \in \mathcal{V}_q}\frac{\dot \alpha_{pq}}{|\mathcal{V}_q|}\mathbf{h}_v\mathbf{W}^k_{V'}
    \end{aligned}
\end{equation}
\cref{eq:approx} elucidates that node $u$ can retrieve information from node $v$ in cluster $p$ by calculating $\frac{\dot \alpha_{pq}}{|\mathcal{V}_q|}\mathbf{h}_v\mathbf{W}^k_{V'}$, where the $\frac{\dot \alpha_{pq}}{|\mathcal{V}_q|}$ represents the approximate attention score between nodes $u$ and $v$.

\section{ Dataset Statistics}
\label{appendix:dataset}
The detailed dataset statistics are listed in \cref{tab:dataset}.
\begin{table}[htb]
\caption{The detailed dataset statistics.}
\label{tab:dataset}
\vskip 0.15in
\begin{center}
\begin{small}
\begin{tabular}{l|c|c|c|c|c}
\toprule
Dataset & \#Nodes & \#Edges & \#Feats & Edge hom & \#Classes \\
\midrule
Cora & 2,708  & 5,429  & 1,433   & 0.83 & 7 \\
CiteSeer & 3,327  & 4,732  & 3,703   & 0.72 & 6 \\
PubMed & 19,717  & 44,338  & 500   & 0.79 & 3 \\
Actor & 7,600  & 26,752  & 931   & 0.22 & 5 \\
Deezer & 28,281  & 92,752  & 31,241   & 0.52 & 2 \\
Ogbn-Arxiv & 169,343  & 1,166,343 &  128   & 0.63 & 40 \\
Ogbn-Products & 2,449,029 & 61,859,140  & 100   & 0.81 & 47 \\
\bottomrule
\end{tabular}
\end{small}
\end{center}
\vskip -0.1in
\end{table}

\section{ Experiments Details}
\label{appendix:details}
In this section, we provide a detailed description of the experimental setup for the empirical results in \cref{experiments}.

\subsection{Training Manner}
We train NAGphormer on all datasets using a mini-batch manner as in the official implementation. For the other baselines and our CoBFormer, we employ a full-batch training method on all datasets except Ogbn-Products, which is too large for all models. Instead, we partition Ogbn-Products into clusters and randomly select several clusters to form a subgraph for each training/evaluation step, ensuring that the size of the combined subgraph remains smaller than our predetermined batch size. We set the batch size to 100,000 for NodeFormer and 150,000 for the other models.

\subsection{CoBFormer}
We implement our CoBFormer using only one BGA module, as SGFormer  has demonstrated that a single global attention layer suffices to achieve optimal expressive capability \cite{sgformer23}. The hyperparameters are selected through grid search within the following search space:
\begin{itemize}
    \item learning rate within $\{ 5e-4, 1e-3, 5e-3, 1e-2, 5e-2 \}$.
    \item GCN layers within $\{ 2, 3 \}$.
    \item weight decay of GCN within $\{ 1e-4, 5e-4, 1e-3, 5e-3, 1e-2 \}$.
    \item weight decay of BGA within $\{ 1e-5, 5e-5, 1e-4, 5e-4, 1e-3 \}$.
    \item number of clusters:
        \begin{itemize}
            \item For Cora, CiteSeer and Actor: $\{ 80, 96, 112, 144 \}$.
            \item For PubMed and Deezer: $\{ 160, 192, 224, 256 \}$.
            \item For Ogbn-Arxiv: $\{ 1536, 2048, 2560, 3072 \}$.
            \item For Ogbn-Products: $\{ 8192, 16384 \}$.
        \end{itemize}
    \item $\alpha$ within: $\{ 0.9, 0.8, 0.7 \}$
    \item $\tau$ within: $\{ 0.9, 0.7, 0.5, 0.3 \}$
\end{itemize}
We fix the dropout ratio of the GCN module at 0.5 and the dropout ratio of the BGA module at 0.1. Layer Normalization is applied to each attention block and FFN block. 

We employ a full-batch training method on all datasets except Ogbn-Products. We partition Ogbn-Products into clusters and randomly select several clusters to form a subgraph for each training/evaluation step, ensuring that the size of the combined subgraph remains smaller than our predetermined batch size.

The Attn-SNR is computed using the intra-cluster attention scores.

\subsection{Baselines}
We implement GCN and GAT by PyG \cite{pyg19} and refer to GammaGL \cite{gammagl}. For the other baselines, we utilize the official repositories. The URLs of these repositories are listed below:
\begin{itemize}
    \item NAGPhormer: https://github.com/JHL-HUST/NAGphormer
    \item NodeFormer: https://github.com/qitianwu/NodeFormer
    \item SGFormer: https://github.com/qitianwu/SGFormer
\end{itemize}

The key hyper-parameters of these baselines are selected from the searching space as follows:
\begin{itemize}
    \item GAT: number of heads within $\{ 1,2,4,8,16 \}$.
    \item NAGPhormer: number of heads within $\{ 1, 2, 4, 8 \}$ and hops within $\{ 3, 5, 7, 10, 15 \}$.
    \item NodeFormer:  rb\_order within $\{ 1, 2, 3 \}$, edge regularization weight within $\{ 0, 1 \}$, number of heads within $\{ 1, 2, 4 \}$ and dropout ratio within $\{ 0, 0.2, 0.3, 0.5 \}$.
    \item SGFormer: $\alpha$ within $\{ 0.5, 0.8 \}$ and dropout ratio within $\{ 0, 0.2, 0.3, 0.5 \}$.
\end{itemize}

\newpage
\section{More Empirical Results}
\subsection{More Visualization Results}
\label{appendix:visual}
\begin{figure*}[ht]
\vskip -0.1in
\begin{center}
\subfigure[]{
\label{fig:Ck3}
\includegraphics[width=0.45\columnwidth]{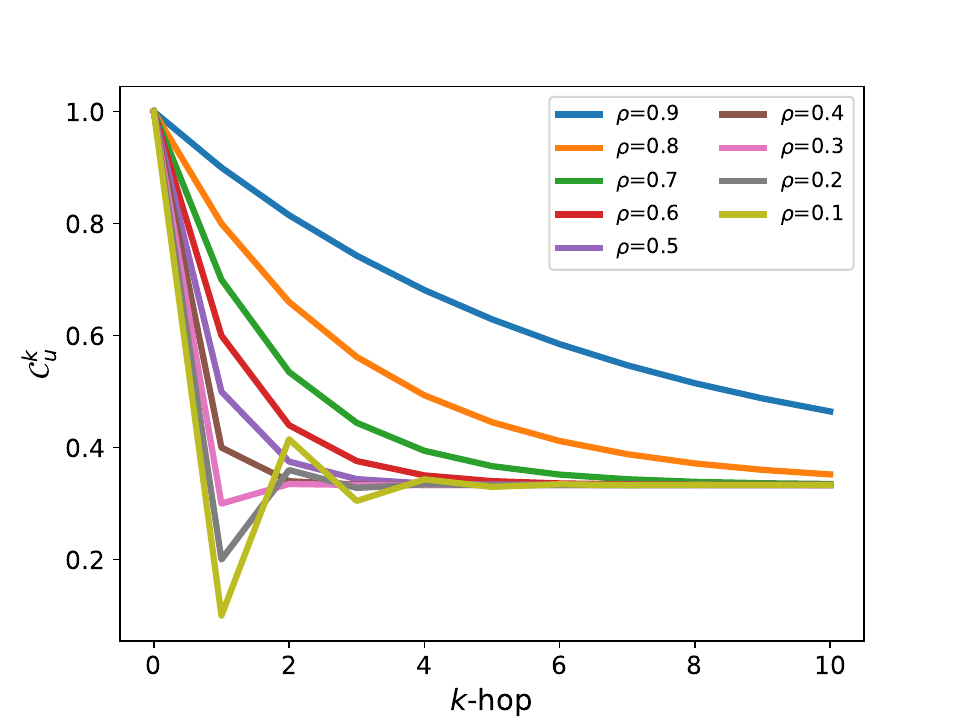}
}\subfigure[]{
\label{fig:Ck4}
\includegraphics[width=0.45\columnwidth]{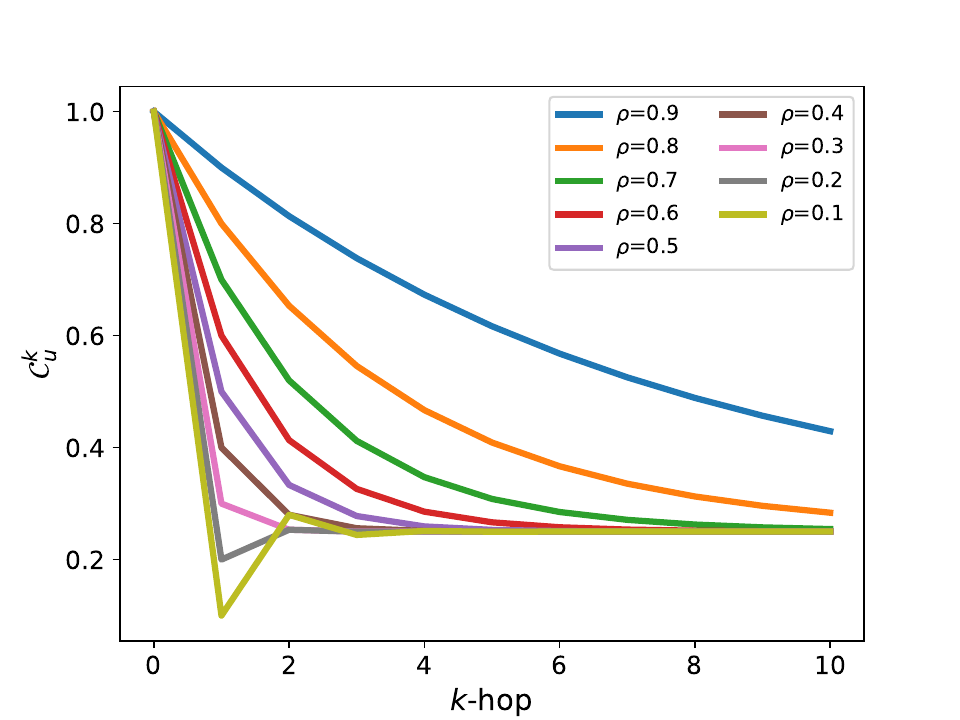}
}
\caption{Visualization results for Equation (6) when $|\mathcal{Y}|=3$ and $|\mathcal{Y}|=4$.}
\end{center}
\vskip -0.1in
\label{fig:ck34}
\end{figure*}

In this section, we will provide additional visualization results for the theoretical variations of $\mathcal{C}_u^k$ when $|\mathcal{Y}|=3$ and $|\mathcal{Y}|=4$, as shown in \cref{fig:Ck3,fig:Ck4}.

\subsection{Results of Significance Test}
\label{appendix:significance}
We conduct a significance test between our method and SGFormer (the best Transformer-based baseline) on four datasets. Specifically, for each dataset, we run each method 50 times and record the accuracy of each run. We then calculate the p-value to assess the statistical significance of the improvements observed with our method. The results for three homophilic graphs are presented in the table below:
\begin{table}[htb]
\caption{Results of significance test on three homophilic graphs.}
\label{tab:significance}
\vskip 0.15in
\begin{center}
\begin{small}
\begin{tabular}{l|c|c}
\toprule
Dataset & p-value(CoB-G vs SGFormer) & p-value(CoB-T vs SGFormer) \\
\midrule
Cora & 1.39e-41  & 3.67e-38 \\
CiteSeer & 5.52e-43  & 1.23e-43 \\
PubMed & 2.95e-47  & 1.49e-52 \\
\bottomrule
\end{tabular}
\end{small}
\end{center}
\vskip -0.1in
\end{table}

The extremely low p-value indicates that our method significantly outperforms the baseline SGFormer in homophilic graphs. Additionally, we have calculated the p-value comparing the accuracies of CoB-T and SGFormer on the Deezer dataset, which is 2.7e-3. This result also suggests a significant improvement of our method in heterophilic graphs.

\section{About Local Module}
\label{appendix:local}
The over-globalizing problem demonstrated in \cref{fig:attn_k_vt,fig:attn_k_node} suggests that a single global attention mechanism tends to allocate most attention scores to distant nodes, resulting in an excessive focus on distant information while neglecting local details. Consequently, integrating a local module to enhance the capture of local information is necessary. Specifically, Gophormer \cite{gophormer21}, ANS-GT \cite{ans-gt22}, NodeFormer \cite{nodeformer22}, and Gapformer \cite{gapformer23} modify the global attention matrix by incorporating the adjacency matrix, thus preserving message passing within the original topological structure. These methods implicitly employ a GNN to capture local information. Coarformer \cite{coarformer22} and SGFormer \cite{sgformer23} explicitly use a GCN as their local module. GOAT \cite{GOAT23} and HSGT \cite{hsgt23} learn local information through neighbor sampling, a technique also utilized by Gophormer and ANS-GT. NAGphormer \cite{NAGphormer22}, another neighbor sampling-based method, employs the aggregated features of the $k$-hop neighborhood as tokens.



\section{A case of Fusion Issue}
\label{appendix:fusion}
\begin{figure}[ht]
\vskip 0.1in
\begin{center}
\centerline{\includegraphics[width=0.6\columnwidth]{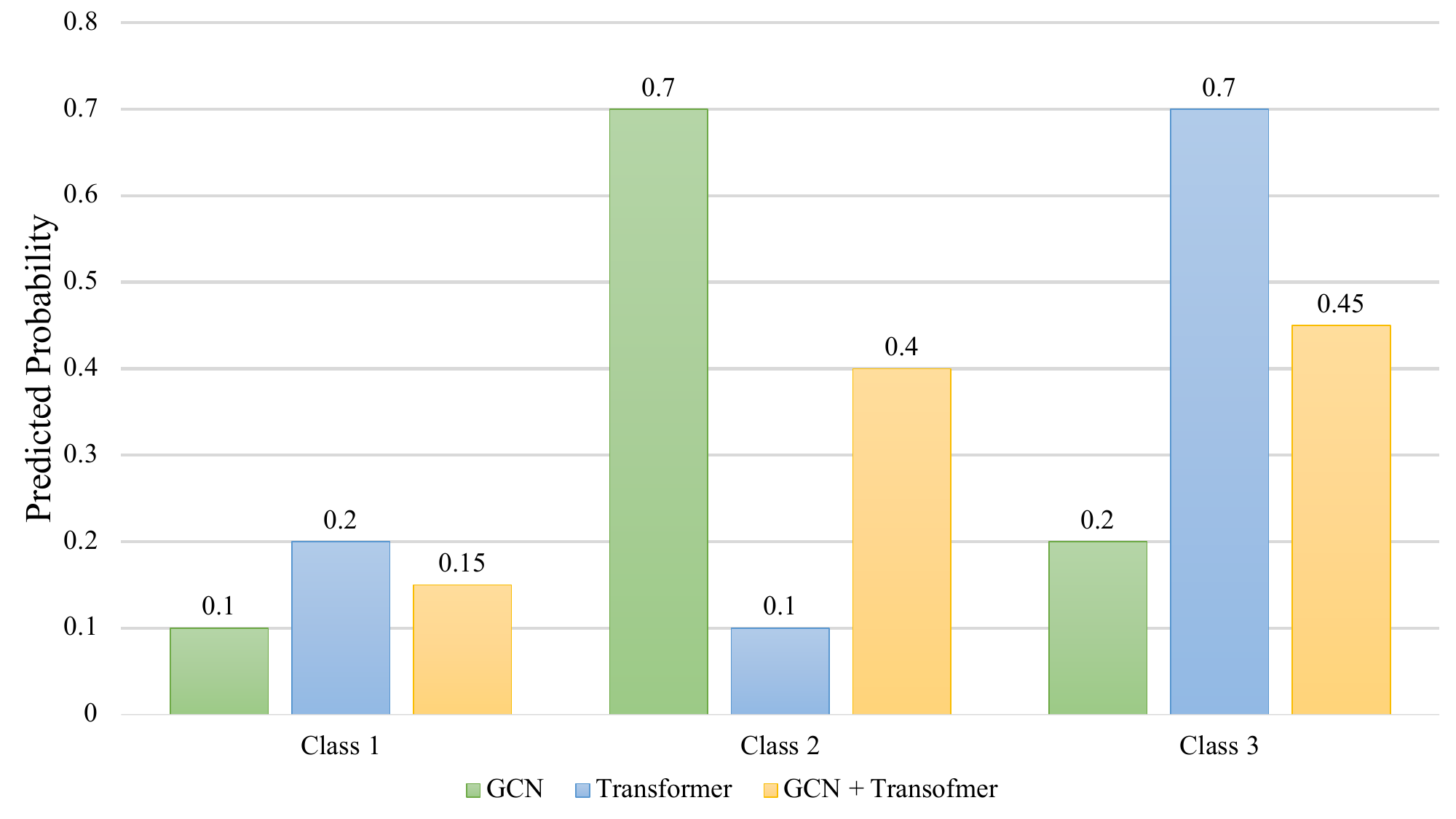}}
\caption{A graphic depiction highlighting the limitations of linear combination is presented. }
\label{fig:TFF}
\end{center}
\vskip -0.2in
\end{figure}

In this section, we will provide a simple example to illustrate the issue we mentioned in \cref{co-train}. As is shown in \cref{fig:TFF}, in this illustrative example, the GCN accurately predicts the node as belonging to the true class 2, whereas the Transformer fails, assigning a high probability to class 3 instead. As a result of the linear combination, the node is incorrectly predicted to be in class 3.

\end{document}